%% file: main.tex
\documentclass{article}

\usepackage{microtype}
\usepackage{graphicx}
\usepackage{subfigure}
\usepackage{booktabs} 

\usepackage{wrapfig}

\usepackage[accepted]{icml2020}

\definecolor{darkblue}{rgb}{0.0, 0.0, 0.55}
\usepackage{amssymb,amsmath,amsthm}

\usepackage[colorlinks,linkcolor=red,anchorcolor=blue,citecolor=darkblue]{hyperref} 
\usepackage[capitalize]{cleveref} 

\usepackage{bm}
\usepackage{multirow}

\newcommand{\bbeta}{{\boldsymbol \beta}}
\newcommand{\bxi}{{\boldsymbol \xi}}
\newcommand{\bW}{{\boldsymbol W}}

\def\hE{\mathbb{E}}
\newcommand{\bphi}{{\boldsymbol \phi}}
\def\cE{{\cal E}}

\newtheorem{theorem}{Theorem}

\newtheorem{lemma}{Lemma}

\newtheorem{assumption}{Assumption}

\edef\oldassumption{\the\numexpr\value{assumption}+1}

\icmltitlerunning{Non-convex Learning via Replica Exchange Stochastic Gradient MCMC}

\begin{document}

\twocolumn[
\icmltitle{Non-convex Learning via Replica Exchange Stochastic Gradient MCMC}






\input{0.authors}

\renewcommand\thesubsection{\Alph{subsection}}
\renewcommand\thesection{\Alph{section}}

\icmlkeywords{Monte Carlo methods, stochastic gradient Langevin dynamics, replica exchange Monte Carlo, Deep neural networks, parallel tempering, ICML}

\vskip 0.3in
]



\printAffiliationsAndNotice{\icmlEqualContribution} 

\input{0.abstract}

\section{Introduction}
\input{1.intro.tex}

\section{Preliminaries}
\input{2.prelim}

\section{Replica Exchange Stochastic Gradient Langevin Dynamics}
\input{3.reSGLD}

\section{Convergence Analysis}
\input{4.convergence}

\section{Experiments}
\input{5.experiments}

\section{Conclusion and Future Work}
\input{6.conclusion}

\section*{Acknowledgements}
\input{7.acknowledge}

\bibliography{mybib}
\bibliographystyle{plainnat}

\newpage
\appendix
\addcontentsline{toc}{section}{Appendices}
\renewcommand{\thesection}{\Alph{section}}
\onecolumn
\input{supp.tex}

\end{document}

%% file: 0.authors.tex
\icmlsetsymbol{equal}{*}

\begin{icmlauthorlist}
\icmlauthor{Wei Deng}{pu}
\icmlauthor{Qi Feng}{equal,usc}
\icmlauthor{Liyao Gao}{equal,pu}
\icmlauthor{Faming Liang}{pu}
\icmlauthor{Guang Lin}{pu}
\end{icmlauthorlist}

\icmlaffiliation{pu}{Purdue University, West Lafayette, IN, USA.}
\icmlaffiliation{usc}{University of Southern California, Los Angeles, CA, USA.}



\icmlcorrespondingauthor{Wei Deng}{weideng056@gmail.com}
\icmlcorrespondingauthor{Guang Lin}{guanglin@purdue.edu}
\icmlcorrespondingauthor{Faming Liang}{fmliang@purdue.edu}

%% file: 0.abstract.tex
\begin{abstract}
Replica exchange Monte Carlo (reMC), also known as parallel tempering, is an important technique for accelerating the convergence of the conventional Markov Chain Monte Carlo (MCMC) algorithms. However, such a method requires the evaluation of the energy function based on the full dataset and is not scalable to big data. The na\"ive implementation of reMC in mini-batch settings introduces large biases, which cannot be directly extended to the stochastic gradient MCMC (SGMCMC), the standard sampling method for simulating from deep neural networks (DNNs). In this paper, we propose an adaptive replica exchange SGMCMC (reSGMCMC) to automatically correct the bias and study the corresponding properties. The analysis implies an acceleration-accuracy trade-off in the numerical discretization of a Markov jump process in a stochastic environment. Empirically, we test the algorithm through extensive experiments on various setups and obtain the state-of-the-art results on CIFAR10, CIFAR100, and SVHN in both supervised learning and semi-supervised learning tasks.
\end{abstract}

%% file: 1.intro.tex
The increasing concern for AI safety problems draws our attention to MCMC, which is known for its asymptotic uncertainty quantification \citep{Chen15, Teh16}, and guarantees in non-convex optimizations \citep{Yuchen17, Maxim17}. Traditional MCMC methods have achieved tremendous success. However, the efficient sampling algorithm in DNNs was not well studied until the invention of stochastic gradient Langevin dynamics (SGLD) \citep{Welling11}, which scales up the computation in DNNs by injecting noises to stochastic gradients. Since then, various high-order SGMCMC algorithms have been proposed, which incorporate strategies such as Hamiltonian dynamics \citep{Chen14, yian2015,Ding14}, Hessian approximation \citep{Li16, Simsekli2016}, and high-order numerical schemes \citep{Chen15, Li19} to improve the convergence.

In addition to the high-order algorithms, we can also follow traditional MCMC algorithms combined with simulated annealing \citep{Kirkpatrick83optimizationby}, simulated tempering \citep{ST}, dynamical weighting \citep{wong97} or replica exchange Monte Carlo \citep{PhysRevLett86, parallel_tempering05}. Among these advancements, simulated annealing SGMCMC \citep{Mangoubi18} and simulated tempering SGMCMC \citep{Holden18} show how dynamical temperatures speed up the convergence. However, simulated annealing is very sensitive to the fast-decaying temperatures, and simulated tempering requires a lot on the approximation of the normalizing constant. For the latter, the replica exchange Monte Carlo is easier to analyze and implement and is suitable for parallelism. Specifically, the replica exchange Langevin diffusion utilizes multiple diffusion processes with different temperatures and proposes to swap the processes while training. Intuitively, the high-temperature process acts as a bridge to connect the various modes. As such, the acceleration effect can be theoretically quantified \citep{Paul12, chen2018accelerating}. However, despite these advantages, a proper replica exchange SGMCMC (reSGMCMC) has long been missing in the deep learning community.

A bottleneck that hinders the development of reSGMCMC is the na\"ive extension of the acceptance-rejection criterion that fails in mini-batch settings. Various attempts \citep{talldata17, Anoop14}
were proposed to solve this issue. 
However, they introduce biases even with the ideal normality assumption on the noise. Some unbiased estimators \citep{PhysRevLett85, Alexandros06} have ever been presented, but the large variance leads to inefficient inference. To remove the bias while maintaining efficiency, \citet{penalty_swap99} proposed a corrected criterion under normality assumptions, and \citet{Daniel17, Matias19} further analyzed the model errors with the asymptotic normality assumptions. However, the above algorithms fail when the required corrections are time-varying and much larger than the energies as shown in Fig.\ref{cifar_biases}(a-b). Consequently, an effective algorithm with the potential to adaptively estimate the corrections and balance between acceleration and accuracy is in great demand.

In this paper, we propose an adaptive replica exchange SGMCMC algorithm via stochastic approximation (SA) \citep{Robbins51, Liang07, deng2019}, a standard method in adaptive sampling to estimate the latent variable: the unknown correction. The adaptive algorithm not only shows the asymptotic convergence in standard scenarios but also gives a good estimate when the corrections are time-varying and excessively large. We theoretically analyze the discretization error for reSGMCMC in mini-batch settings and show the accelerated convergence in 2-Wasserstein distance. Such analysis sheds light on the use of biased estimates of unknown corrections to obtain a trade-off between acceleration and accuracy. In summary, this algorithm has three main contributions:

1. We propose a novel reSGMCMC to speed up the computations of SGMCMC in DNNs with theoretical guarantees. The theory shows the potential of using biased corrections and a large batch size to obtain better performance.

2. We identify the problem of time-varying corrections in DNNs and propose to adaptively estimate the time-varying corrections, with potential extension to a variety of time-series prediction techniques.

3. We test the algorithm through extensive experiments using various models. It achieves the state-of-the-art results in both supervised learning and semi-supervised learning tasks on CIFAR10, CIFAR100, and SVHN datasets.

%% file: 2.prelim.tex
A standard sampling algorithm is the Langevin diffusion, which is a stochastic differential equation (SDE) as follows:
\begin{equation}
\label{sde_1}
    d \bbeta_t^{(1)} = - \nabla U(\bbeta_t^{(1)}) dt+\sqrt{2\tau_1} d\bW_t^{(1)},
\end{equation}
where $\bbeta_t^{(1)}\in\mathbb{R}^d$, $U(\cdot)$ is the energy function, $\bW_t^{(1)}\in\mathbb{R}^d$ is the Brownian motion, and $\tau_1>0$ is the temperature. 

Under mild growth conditions on $U$, the Langevin diffusion $\{\bbeta_t^{(1)}\}_{t\geq 0}$ converges to the unique invariant Gibbs distribution $\pi_{\tau_1}(\bbeta^{(1)})\propto \exp(-\frac{U(\bbeta^{(1)})}{\tau_1})$, where the temperature $\tau_1$ is crucial for both optimization and sampling of the non-convex energy function $U$. On the one hand, a high-temperature $\tau_1$ achieves the \emph{exploration} effect: the convergence to the flattened Gibbs distribution of the whole domain is greatly facilitated. However, the flattened distribution is less concentrated around the global optima 
\citep{Maxim17}, and the geometric connection to the global minimum is affected \citep{Yuchen17}.  
On the other hand, a low-temperature $\tau_1$ leads to the \emph{exploitation} effect: the solutions explore the local geometry rapidly, but they are more likely to get trapped in local optima, leading to a slow convergence in both optimization and sampling. Therefore, the potential of using a fixed temperature is quite limited.

\begin{wrapfigure}{r}{0.21\textwidth}
   \begin{center}
   \vskip -0.2in
     \includegraphics[width=0.21\textwidth]{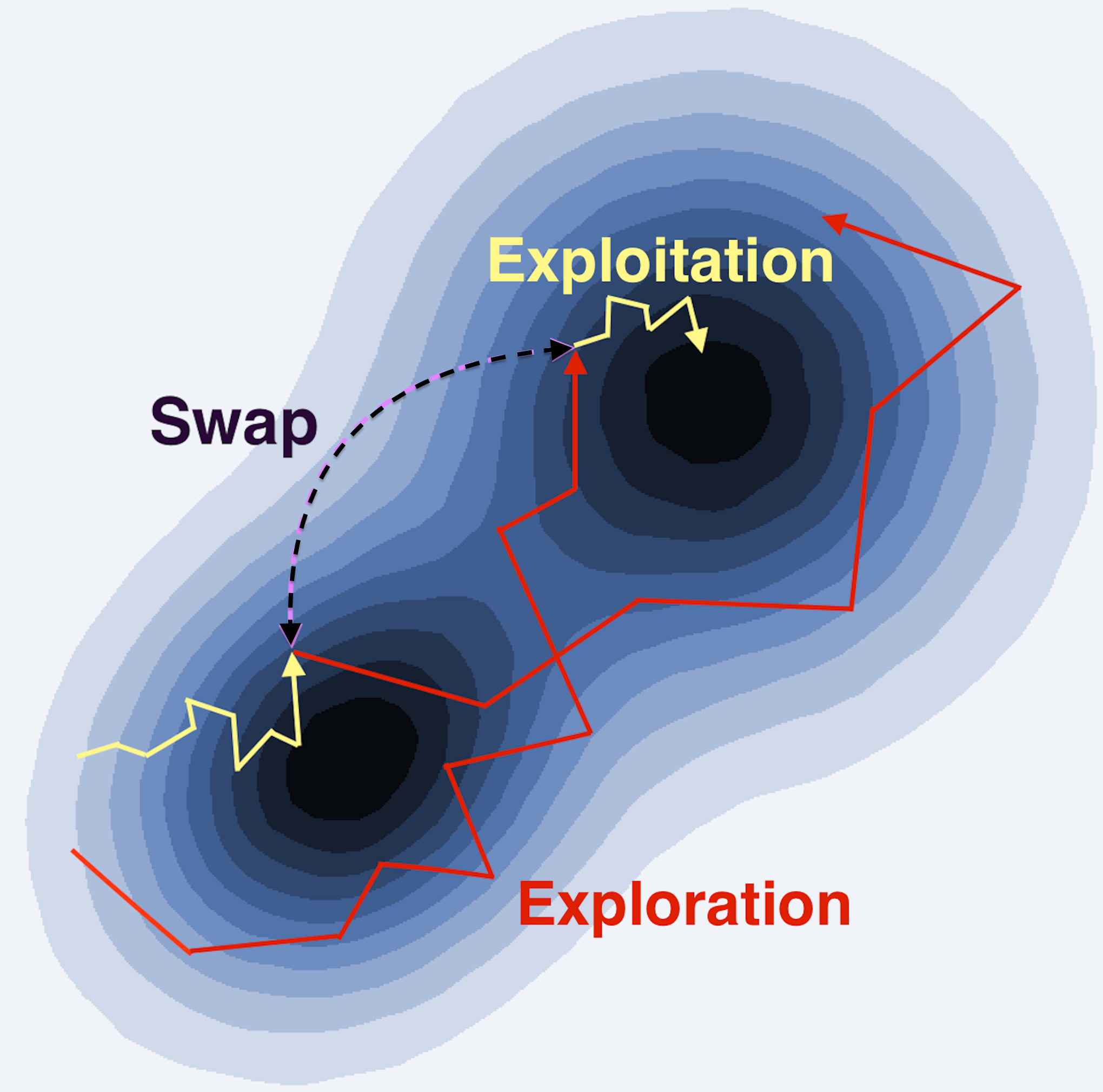}
   \end{center}
   \vskip -0.2in
   \caption{Paths of reLD.}
   \label{path_demo}
\end{wrapfigure}

A powerful algorithm called replica exchange Langevin diffusion (reLD), also known as parallel tempering Langevin diffusion, has been proposed to accelerate the convergence of the SDE as shown in Fig.\ref{path_demo}. 
reLD proposes to simulate a high-temperature particle for \emph{exploration} and a low-temperature particle for \emph{exploitation} and allows them to swap simultaneously. Now consider the following coupled processes with a higher temperature $\tau_2>\tau_1$ and $\bW^{(2)}$ independent of $\bW^{(1)}$:
\begin{equation}
\label{sde_2_couple}
\begin{split}
    d \bbeta_t^{(1)} &= - \nabla U(\bbeta_t^{(1)}) dt+\sqrt{2\tau_1} d\bW_t^{(1)}\\
    d \bbeta_t^{(2)} &= - \nabla U(\bbeta_t^{(2)}) dt+\sqrt{2\tau_2} d\bW_t^{(2)}.\\
\end{split}
\end{equation}
Eq.(\ref{sde_2_couple}) converges to the invariant distribution with density 
\begin{equation}
\label{pt_density_main}
\begin{split}
    \pi(\bbeta^{(1)}, \bbeta^{(2)})\propto e^{-\frac{U(\bbeta^{(1)})}{\tau_1}-\frac{U(\bbeta^{(2)})}{\tau_2}}.
\end{split}
\end{equation}
By allowing the two particles to swap, the positions are likely to change from $(\bbeta_t^{(1)},\bbeta_t^{(2)})$ to $(\bbeta_{t+dt}^{(2)},\bbeta_{t+dt}^{(1)})$ with a swapping rate $r (1\wedge S(\bbeta_t^{(1)},\bbeta_t^{(2)}))dt$, where the constant $r\geq 0$ is the swapping intensity, and $S(\cdot, \cdot)$ satisfies
\begin{equation}
\begin{split}
    S(\bbeta_t^{(1)}, \bbeta_t^{(2)}):=e^{ \left(\frac{1}{\tau_1}-\frac{1}{\tau_2}\right)\left(U(\bbeta_t^{(1)})-U(\bbeta_t^{(2)})\right)}.
\end{split}
\end{equation}
In such a case, reLD is a Markov jump process, which is reversible \citep{chen2018accelerating} and leads to the same invariant distribution ($\ref{pt_density_main}$).


%% file: 3.reSGLD.tex
The wide adoption of the replica exchange Monte Carlo in traditional MCMC algorithms motivates us to design replica exchange stochastic gradient Langevin dynamics for DNNs, but the straightforward extension of reLD to replica exchange stochastic gradient Langevin dynamics is highly non-trivial \citep{Chen14, yian2015, Simsekli2016}. In this section, we will first show that na\"ive extensions of replica exchange Monte Carlo to SGLD (na\"ive reSGLD) lead to large biases. Afterward, we will present an adaptive replica exchange stochastic gradient Langevin dynamics (reSGLD) that will automatically adjust the bias and yield a good approximation to the correct distribution.

\subsection{Na\"ive reSGLD}

We denote the entire data by $\mathcal{D}=\{\bm{\mathrm{d}}_i\}_{i=1}^N$, where $\bm{\mathrm{d}}_i$ is a data point. Given the model parameter $\widetilde \bbeta$, we consider the following energy function (negative log-posterior)
\begin{equation}
L(\widetilde \bbeta)= -\log p(\widetilde \bbeta) - \sum_{i=1}^N \log P(\bm{\mathrm{d}}_i|\widetilde \bbeta).
\end{equation}
where $p(\widetilde \bbeta)$ is a proper prior and $\sum_{i=1}^N \log P(\bm{\mathrm{d}}_i|\widetilde \bbeta)$ is the complete data log-likelihood. When the number of data points $N$ is large, it is expensive to evaluate $L(\widetilde \bbeta)$ directly. Instead, we propose to approximate the energy function with a mini-batch of data $\mathcal{B}=\{\bm{\mathrm{d}}_{s_i}\}_{i=1}^n$, where $s_i\in \{1, 2, ..., N\}$. We can easily check that if $\mathcal{B}$ is sampled randomly with or without replacement, we obtain the following unbiased estimator of the energy function
\begin{equation}
\widetilde L(\widetilde \bbeta) = -\log p(\widetilde \bbeta)-\frac{N}{n}\sum_{i=1}^n \log P(\bm{\mathrm{d}}_{s_i}|\widetilde \bbeta).
\end{equation}
Let $\widetilde \bbeta_k$ denote the estimate of $\widetilde \bbeta$ at the $k$-th iteration. Next, SGLD proposes the following iterations:
\begin{equation}
\begin{split}
&\widetilde \bbeta_{k+1}=\widetilde \bbeta_k - \eta_k \nabla \widetilde{L}(\widetilde \bbeta_k)+\sqrt{2\eta_k\tau_1}\bxi_k,\\
\end{split}
\end{equation}
where $\eta_k$ is the learning rate, the stochastic gradient $\nabla \widetilde{L}(\widetilde \bbeta_k)$ is the unbiased estimator of the exact gradient $\nabla L(\widetilde \bbeta_k)$, $\bxi$ is a standard $d$-dimensional Gaussian vector with mean $\bm{0}$ and identity covariance matrix. It is known that SGLD asymptotically converges to a unique invariant distribution $\pi(\widetilde \bbeta)\propto \exp\left(-L(\widetilde \bbeta)/\tau_1\right)$ \citep{Teh16} as $\eta_k\rightarrow 0$. If we simply replace gradients with stochastic gradients in the replica exchange gradient Langevin dynamics, we have
\begin{equation}
\label{naive_resgld}
\begin{split}
    \widetilde \bbeta^{(1)}_{k+1} &= \widetilde \bbeta^{(1)}_{k}- \eta_k \nabla \widetilde L(\widetilde \bbeta^{(1)}_k)+\sqrt{2\eta_k\tau_1} \bxi_k^{(1)}\\
    \widetilde \bbeta^{(2)}_{k+1} &= \widetilde \bbeta^{(2)}_{k} - \eta_k\nabla \widetilde L(\widetilde \bbeta^{(2)}_k)+\sqrt{2\eta_k\tau_2} \bxi_k^{(2)}.\\
\end{split}
\end{equation}
Furthermore, we swap the Markov chains in (\ref{naive_resgld}) with the na\"ive stochastic swapping rate $r (1\wedge\mathbb{S}(\widetilde \bbeta_{k+1}^{(1)}, \widetilde \bbeta_{k+1}^{(2)}))\eta_k$\footnote[4]{In the implementations, we fix $r\eta_k=1$ by default.}:
\begin{equation}
\label{swap_naive}
\begin{split}
    &\mathbb{S}(\widetilde \bbeta_{k+1}^{(1)}, \widetilde \bbeta_{k+1}^{(2)})=e^{ \left(\frac{1}{\tau_1}-\frac{1}{\tau_2}\right)\left(\widetilde L(\widetilde \bbeta_{k+1}^{(1)})-\widetilde L(\widetilde \bbeta_{k+1}^{(2)})\right)}.\\
\end{split}
\end{equation}
Apparently, the unbiased estimators $\widetilde L(\widetilde \bbeta_{k+1}^{(1)})$ and $\widetilde L(\widetilde \bbeta_{k+1}^{(2)})$ in $\mathbb{S}(\widetilde \bbeta_{k+1}^{(1)}, \widetilde \bbeta_{k+1}^{(2)})$ do not provide an unbiased estimator of $S(\widetilde \bbeta_{k+1}^{(1)}, \widetilde \bbeta_{k+1}^{(2)})$ after a non-linear transformation as shown in (\ref{swap_naive}), which leads to a large bias.



\subsection{Replica Exchange Stochastic Gradient Langevin Dynamics with Correction}
A viable MCMC algorithm requires the approximately unbiased estimators of the swapping rates to ``satisfy'' the detailed balance property \citep{penalty_swap99, Andrieu09, Geoff12} and the weak solution of a Markov jump process with unbiased stochastic coefficients has also been studied in \citet{Gyongy86, Amel}. When we make normality assumption on the stochastic energy $\widetilde L(\bbeta)\sim\mathcal{N}( L(\bbeta), \sigma^2)$, it follows 
\begin{equation}
\label{diff_L}
    \widetilde L(\widetilde \bbeta^{(1)})-\widetilde L(\widetilde \bbeta^{(2)})=
    L(\widetilde \bbeta^{(1)})-L(\widetilde \bbeta^{(2)}) + \sqrt{2}\sigma W_1, \\
\end{equation}
where $W_1$ follows the standard normal distribution and can be viewed as a Brownian motion at $t=1$. Consider the evolution of the stochastic swapping rate $\{\widetilde S_t\}_{t\in[0,1]}$ in each swap as a geometric Brownian motion:
\begin{equation}
\small
\begin{split}
\label{analytic}
    \widetilde S_t&=e^{\left(\frac{1}{\tau_1}-\frac{1}{\tau_2}\right)\left(\widetilde L(\widetilde \bbeta^{(1)})-\widetilde L(\widetilde \bbeta^{(2)})-\left(\frac{1}{\tau_1}-\frac{1}{\tau_2}\right)\sigma^2 t\right)}\\
    &=e^{\left(\frac{1}{\tau_1}-\frac{1}{\tau_2}\right)\left(L(\widetilde \bbeta^{(1)})-L(\widetilde \bbeta^{(2)})-\left(\frac{1}{\tau_1}-\frac{1}{\tau_2}\right)\sigma^2 t+\sqrt{2}\sigma W_t\right)}.\\
\end{split}
\end{equation}
Set $\tau_{\delta}=\frac{1}{\tau_1}-\frac{1}{\tau_2}$ and take the partial derivatives of $\widetilde S_t$
\begin{equation*}
\begin{split}
     &\frac{d \widetilde S_t}{d t}=-\tau_{\delta}^2\sigma^2\widetilde S_t,\  \frac{d \widetilde S_t}{dW_t}=\sqrt{2}\tau_{\delta}\sigma \widetilde S_t,\frac{d^2 \widetilde S_t}{d W_t^2}=2\tau_{\delta}^2\sigma^2 \widetilde S_t.
\end{split}
\end{equation*}
It\^{o}'s lemma shows that
\begin{equation*}
\begin{split}
    d \widetilde S_t&=\left(\frac{d\widetilde S_t}{dt}+\frac{1}{2}\frac{d^2 \widetilde S_t}{dW_t^2}\right)dt + \frac{d\widetilde S_t}{dW_t}dW_t= \sqrt{2}\tau_{\delta}\sigma \widetilde S_t d W_t.
\end{split}
\end{equation*}
Notice that $\{\widetilde S_t\}_{t\in[0,1]}$ is a Martingale and yields the same expectation for $\forall t\in [0, 1]$. By fixing $t=1$ in (\ref{analytic}), we have
\begin{equation}
\begin{split}
\label{sto_acceptance}
    \widetilde S_1&=e^{ \left(\frac{1}{\tau_1}-\frac{1}{\tau_2}\right)\left( \widetilde L(\widetilde \bbeta^{(1)})- \widetilde L(\widetilde \bbeta^{(2)})-\left(\frac{1}{\tau_1}-\frac{1}{\tau_2}\right)\sigma^2\right)},\\
\end{split}
\end{equation}
where the stochastic swapping rate $\widetilde S_1$ is an unbiased estimator of $\widetilde S_0=e^{\left(\frac{1}{\tau_1}-\frac{1}{\tau_2}\right)\left(L(\widetilde \bbeta^{(1)})-L(\widetilde \bbeta^{(2)})\right)}$, and the correction term $\left(\frac{1}{\tau_1}-\frac{1}{\tau_2}\right)\sigma^2$ aims to remove the bias from the swaps. 

An advantage of interpreting the correction term from the perspective of geometric Brownian motion is that we may extend it to geometric L\'{e}vy process \citep{David04}, which is more suitable for the heavy-tailed energy noise \citep{Simsekli2019b}. Admittedly, the estimation of the tail-index of extreme-value distributions and the correction under L\'{e}vy process go beyond the scope of this paper, so we leave it for future works.

\subsection{Adaptive Replica Exchange Stochastic Gradient Langevin Dynamics}

In reality, the exact variance $\sigma^2$ is hardly known and subject to estimation. The normality assumption may be violated and even no longer time-independent.

\subsubsection{Fixed Variance $\sigma^2$}
We use stochastic approximation (SA) to adaptively estimate the unknown variance while sampling from the posterior. In each SA step, we obtain an unbiased sample variance $\tilde \sigma^2$ for the true $\sigma^2$ and update the adaptive estimate $\hat \sigma_{m+1}^2$ through
\begin{equation}
    \hat \sigma^2_{m+1} = (1-\gamma_m)\hat \sigma^2_{m}+\gamma_m \tilde \sigma^2_{m+1},
\end{equation}
where $\gamma_m$ is the smoothing factor at the $m$-th SA step. The SA step is updated less frequently than the standard sampling to reduce the computational cost. When the normality assumption holds, we notice that $\hat \sigma_{m}^2=\sum_{i=1}^m \tilde \sigma_{i}^2/m$ when $\gamma_m=\frac{1}{m}$. Following central limit theorem (CLT), we have that $\hat \sigma_m^2-\sigma^2 =\mathcal{O}(\frac{1}{m})$. 
Inspired by theorem 2 from \citet{Chen15}, we expect that the weak convergence of the adaptive sampling algorithm holds since the bias decreases sufficiently fast ($\frac{1}{m}\sum_{l=1}^{m} \mathcal{O}(\frac{1}{l})\rightarrow 0$ as $m\rightarrow \infty$).

In practice, the normality assumption is likely to be violated when we use a small batch size $n$, but the unknown distribution asymptotically approximates the normal distribution as $n\rightarrow \infty$ and yield a bias $\mathcal{O}(\frac{1}{n})$ in each SA step. Besides, the mini-batch setting usually introduces a very large noise on the estimator of the energy function, which requires a large correction term and leads to \emph{almost-zero swapping rates}. 

To handle this issue, we introduce a correction factor $F$ to reduce the correction term from $\left(\frac{1}{\tau_1}-\frac{1}{\tau_2}\right)\hat \sigma^2$ to $\frac{\left(\frac{1}{\tau_1}-\frac{1}{\tau_2}\right)\hat \sigma^2}{F}$. We note that a large $F>1$ introduces some bias, but may significantly increase the acceleration effect, giving rise to an acceleration-accuracy trade-off in finite time. Now, we show the algorithm in Alg.\ref{alg}. In addition to simulations of multi-modal distributions, our algorithm can be also combined with simulated annealing \citep{PT_SA, OPT_SA} 
to accelerate the non-convex optimization and increase the hitting probability to the global optima \cite{Mangoubi18}.

\begin{algorithm}[tb]
   \caption{Adaptive Replica Exchange Stochastic Gradient Langevin Dynamics Algorithm. For \emph{sampling purposes}, we fix the temperatures $\tau_1$ and $\tau_2$; for \emph{optimization purposes}, we keep annealing $\tau_1$ and $\tau_2$ during each epoch. Empirically, a larger $\gamma_m$ tracks the dynamics better but is less robust. The intensity $r$ and $\eta$ are omitted in the corrected swaps.}
   \label{alg}
\begin{algorithmic}
\REPEAT
   \STATE{\textbf{Sampling Step}}
   \begin{equation*}
   \small
       \begin{split}
           \widetilde \bbeta^{(1)}_{k+1} &= \widetilde \bbeta^{(1)}_{k}- \eta_k^{(1)} \nabla \widetilde L(\widetilde \bbeta^{(1)}_k)+\sqrt{2\eta_k^{(1)}\tau_1} \bxi_k^{(1)}\\
    \widetilde \bbeta^{(2)}_{k+1} &= \widetilde \bbeta^{(2)}_{k} - \eta_k^{(2)}\nabla \widetilde L(\widetilde \bbeta^{(2)}_k)+\sqrt{2\eta_k^{(2)}\tau_2} \bxi_k^{(2)},\\
       \end{split}
   \end{equation*}
   \STATE{\textbf{SA Step}}
   \STATE{Obtain an unbiased estimate $\tilde \sigma_{m+1}^2$ for $\sigma^2$.}
   \begin{equation*}
   \begin{split}
       &\hat \sigma^2_{m+1} = (1-\gamma_m)\hat \sigma^2_{m}+\gamma_m \tilde \sigma^2_{m+1},\\
   \end{split}
   \end{equation*}
   \STATE{\textbf{Swapping Step}}
   \STATE{Generate a uniform random number $u\in [0,1]$.}
   \begin{equation*}
       \textstyle \hat S_1=e^{ \left(\frac{1}{\tau_1}-\frac{1}{\tau_2}\right)\left( \widetilde L(\widetilde \bbeta_{k+1}^{(1)})- \widetilde L(\widetilde \bbeta_{k+1}^{(2)})-\frac{\left(\frac{1}{\tau_1}-\frac{1}{\tau_2}\right)\hat \sigma^2_{m+1}}{F}\right)}.
   \end{equation*}
   \IF{$u<\hat S_1$}
   \STATE Swap $\widetilde \bbeta_{k+1}^{(1)}$ and $\widetilde \bbeta_{k+1}^{(2)}$.
   \ENDIF
   \UNTIL{$k=k_{\max}$}
    \vskip -1 in
\end{algorithmic}
\end{algorithm}

\subsubsection{Time-varying Variance $\sigma^2$}
In practice, the variance $\sigma^2$ usually varies with time, resulting in time-varying corrections. For example, in the optimization of residual networks on CIFAR10 and CIFAR100 datasets, we notice from Fig.\ref{cifar_biases}(a-b) that the corrections are time-varying. As such, we cannot use a fixed correction anymore to deal with the bias. The treatment for the time-varying corrections includes standard methods for time-series data, and a complete recipe for modeling the data goes beyond our scope. We still adopt the method of stochastic approximation and choose a fixed smoothing factor $\gamma$ so that
\begin{equation}
\label{smoothing}
    \hat \sigma^2_{m+1} = (1-\gamma)\hat \sigma^2_{m}+\gamma \tilde \sigma^2_{m+1}.
\end{equation}

Such a method resembles the simple exponential smoothing and acts as robust filters to remove high-frequency noise. It can be viewed as a special case of autoregressive integrated moving average (ARIMA) (0,1,1) model but often outperforms the ARIMA equivalents because it is less sensitive to the model selection error \citep{john66}. From the regression perspective, this method can be viewed as a zero-degree
local polynomial kernel model \cite{Gijbels99}, which is robust to distributional
assumptions.

%% file: 4.convergence.tex
We theoretically analyze the acceleration effect and the accuracy of reSGLD in terms of 2-Wasserstein distance between the Borel probability measures $\mu$ and $\nu$ on $\mathbb{R}^d$ 
\begin{equation*}
    \mathcal{W}_2(\mu, \nu):=\inf_{\Gamma\in \text{Couplings}(\mu, \nu)}\sqrt{\int\|\bbeta_{\mu}-\bbeta_{\nu}\|^2 d \Gamma(\bbeta_{\mu}, \bbeta_{\nu})},
\end{equation*}
where $\|\cdot\|$ is the Euclidean norm, and the infimum is taken over all joint distributions $\Gamma(\bbeta_{\mu}, \bbeta_{\nu})$ with $\mu$ and $\nu$ being its marginal distributions.

Our analysis begins with the fact that reSGLD in Algorithm.\ref{alg} tracks the replica exchange Langevin diffusion (\ref{sde_2_couple}). For ease of analysis, we consider a fixed learning rate $\eta$ for both chains. reSGLD can be viewed as a special discretization of the continuous-time Markov jump process. In particular, it differs from the standard discretization of the continuous-time Langevin algorithms \cite{chen2018accelerating, yin_zhu_10, Maxim17,Issei14} in that we need to consider the discretization of the Markov jump process in a stochastic environment. To handle this issue, we follow \citet{Paul12} and view the swaps of positions as swaps of the temperatures, which have been proven equivalent in distribution. 
\begin{lemma}[Discretization Error]
\label{th:1}
Given the smoothness and dissipativity assumptions in the appendix, and a small learning rate $\eta$, we have that
\begin{equation*}
    \begin{split}
&    \scriptstyle \hE[\sup_{0\le t\le T}\|\bbeta_t-\widetilde \bbeta^{\eta}_t||^2] \le \mathcal{\tilde O}(\eta+\max_i\hE[\|\bphi_i\|^2]+ \max_{i}\sqrt{\hE\left[|\psi_{i}|^2\right]}),\\
    \end{split}
\end{equation*}
where $\widetilde \bbeta^{\eta}_t$ is the continuous-time interpolation for reSGLD, $\bphi:=\nabla \widetilde U-\nabla U$ is the noise in the stochastic gradient, and $\psi:=\widetilde S-S$ is the noise in the stochastic swapping rate. 
\end{lemma}

Then we quantify the evolution of the 2-Wasserstein distance between $\nu_{t}$ and the invariant distribution $\pi$, where $\nu_t$ is the probability measure associated with reLD at time $t$. The key tool is the exponential decay of entropy when $\pi$ satisfies the log-Sobolev inequality (LSI) \citep{Bakry2014}. To justify LSI, we first verify LSI for reLD without swaps, which is a direct result \citep{Cattiaux2010} given the Lyapunov function criterion and the Poincar\'{e} inequality \citep{chen2018accelerating}. Then we verify LSI for reLD with swaps by analyzing the corresponding Dirichlet form, which is strictly larger than the Dirichlet form associated with reLD without swaps. Finally, the exponential decay of the 2-Wasserstein distance follows from the Otto-Villani theorem \citep{Bakry2014} by connecting 2-Wasserstein distance with the relative entropy.

\begin{lemma}[Accelerated exponential decay of $\mathcal{W}_2$]\label{exponential decay_main}
Under the smoothness and dissipativity assumptions, we have that the replica exchange Langevin diffusion converges exponentially fast to the invariant distribution $\pi$:
\begin{equation}
    \mathcal{W}_2(\nu_t,\pi) \leq  D_0 e^{-k\eta(1+\delta_S)/c_{\text{LS}}},
\end{equation}
where $\delta_{S}:=\inf_{t>0}\frac{\cE_S(\sqrt{\frac{d\nu_t}{d\pi}})}{\cE(\sqrt{\frac{d\nu_t}{d\pi}})}-1$ is the very \textbf{acceleration effect} depending on the swapping rate $S$, $\cE$ and $\cE_S$ are the Dirichlet forms defined in the appendix, $c_{\text{LS}}$ is the constant in the log-Sobolev inequality, $D_0=\sqrt{2c_{\text{LS}}D(\nu_0||\pi)}$.
\end{lemma}{}

Finally, combining the definition of Wasserstein distance and the triangle inequality, we have that 

\begin{theorem}[Convergence of reSGLD]Let the smoothness and dissipativity assumptions hold. For the distribution $\{\mu_{k}\}_{k\ge 0}$ associated with the discrete dynamics $\{\widetilde \bbeta_k\}_{k\ge 1}$, we have the following estimates for $k\in \mathbb N^{+}$:
\begin{equation*}
\begin{split}
    \mathcal{W}_2(\mu_{k}, &\pi) \le  D_0 e^{-k\eta(1+\delta_S)/c_{\text{LS}}}\\
    &+\footnotesize{\mathcal{\tilde O}(\eta^{\frac{1}{2}}+\max_i(\hE[\|\bphi_i\|^2])^{\frac{1}{2}}+ \max_{i}(\hE\left[|\psi_{i}|^2\right]})^{\frac{1}{4}}),\\
\end{split}
\end{equation*}
where $D_0=\sqrt{2c_{\text{LS}}D(\mu_0||\pi)}$, $\delta_{S}:=\min_{i}\frac{\cE_S(\sqrt{\frac{d\mu_i}{d\pi}})}{\cE(\sqrt{\frac{d\mu_i}{d\pi}})}-1$.
\end{theorem}{}
Ideally, we want to boost the acceleration effect $\delta_S$ by using a larger swapping rate $S$ and increase the accuracy by reducing the mean squared errors $\hE[\|\bphi_i\|^2]$ and $\hE[|\psi_i|^2]$. One possible way is to apply a \emph{large enough batch size}, which may be yet inefficient given a large dataset. Another way is to \emph{balance} between acceleration and accuracy by tuning the correction factor $F$. In practice, a larger $F$ leads to a larger acceleration effect and also injects more biases.

%% file: 5.experiments.tex
\subsection{Simulations of Gaussian Mixture Distributions}

In this group of experiments, we evaluate the acceleration effects and the biases for reSGLD on multi-modal distributions based on different assumptions on the estimators for the energy function. As a comparison, we choose SGLD and the na\"ive reSGLD without corrections as baselines. The learning rates $\eta^{(1)}$ and $\eta^{(2)}$ are both set to 0.03, and the temperatures $\tau_1$ and $\tau_2$ are set to 1 and 10, respectively. In particular, SGLD uses the learning rate $\eta^{(1)}$ and the temperature $\tau_1$. We simulate 100,000 samples from each distribution and propose to estimate the correction every 100 iterations. The correction estimator is calculated based on the variance of 10 samples of $\widetilde U_1(x)$. The initial correction is set to 100 and the step size $\gamma_m$ for stochastic approximation is chosen as $\frac{1}{m}$. The correction factor $F$ is 1 in the first two examples.

\begin{figure*}[!ht]
  \centering
  \subfigure[$\scriptstyle \widetilde U_1(x)\sim \mathcal{N}(U_1(x), 2^2)$.]{\includegraphics[scale=0.22]{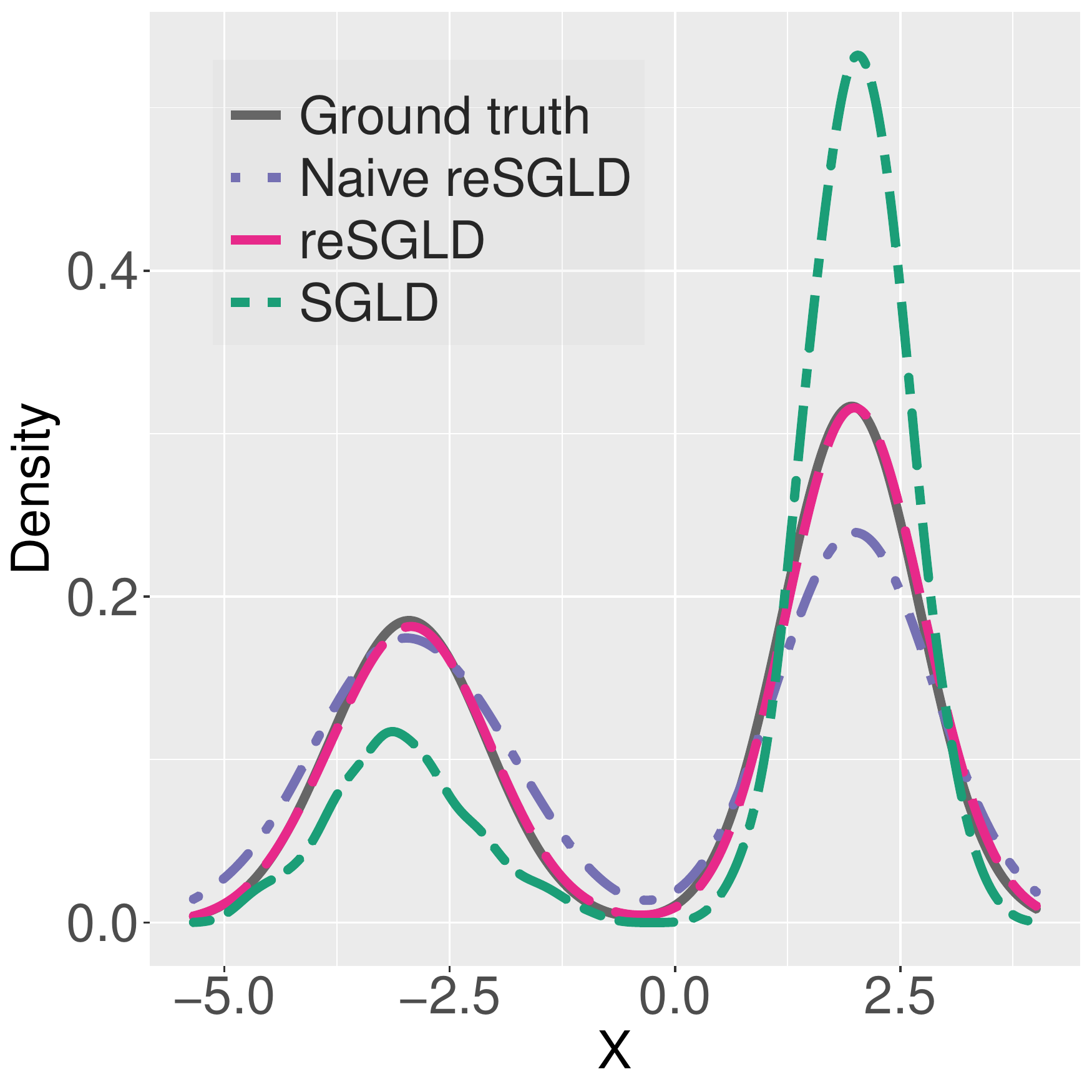}}\label{fig: 22a}\quad
  \subfigure[Convergence evaluation.]{\includegraphics[scale=0.22]{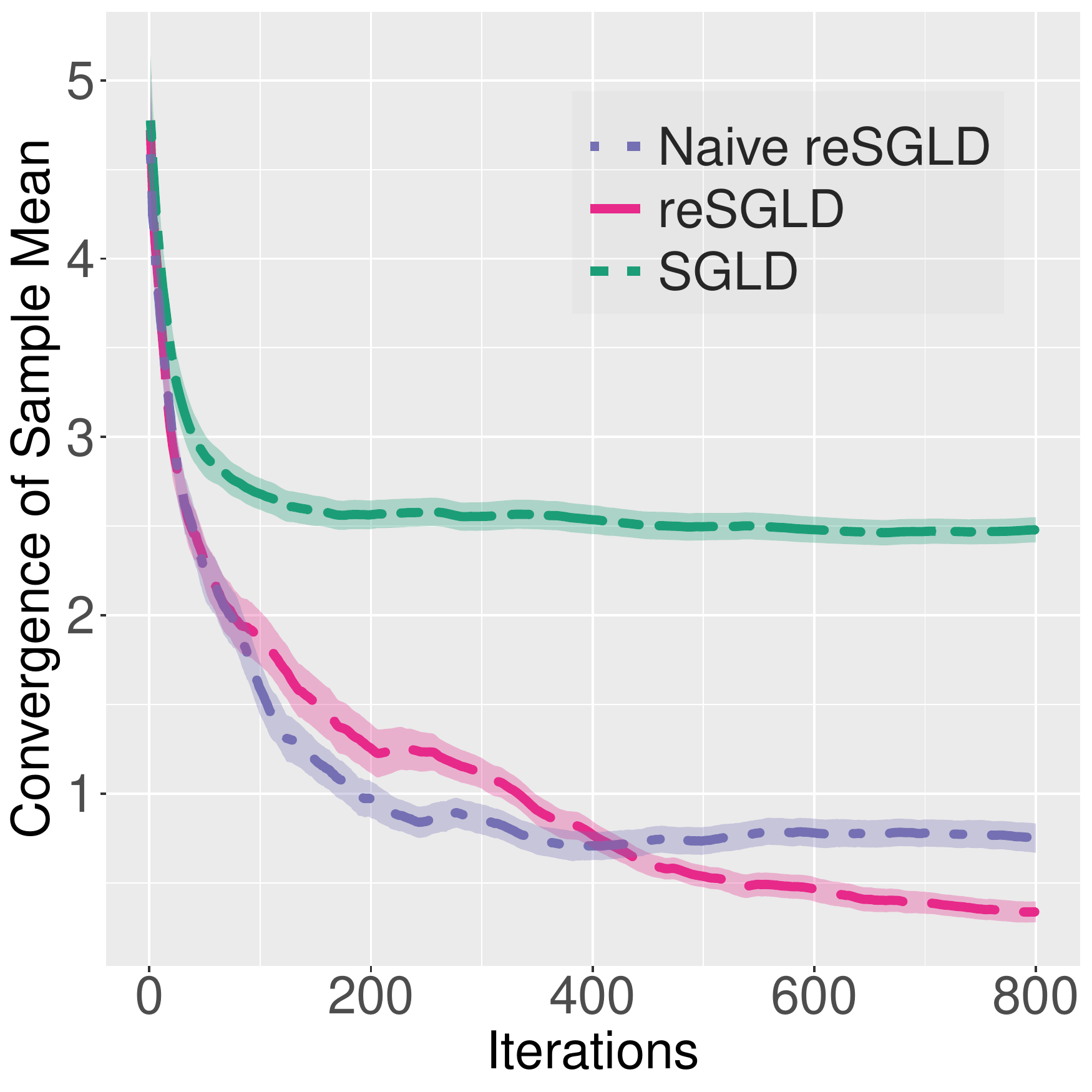}}\label{fig: 22b}\quad
  \subfigure[$\scriptstyle \widetilde U_2(x)\sim U_2(x)+ t(\nu=5)$.]{\includegraphics[scale=0.22]{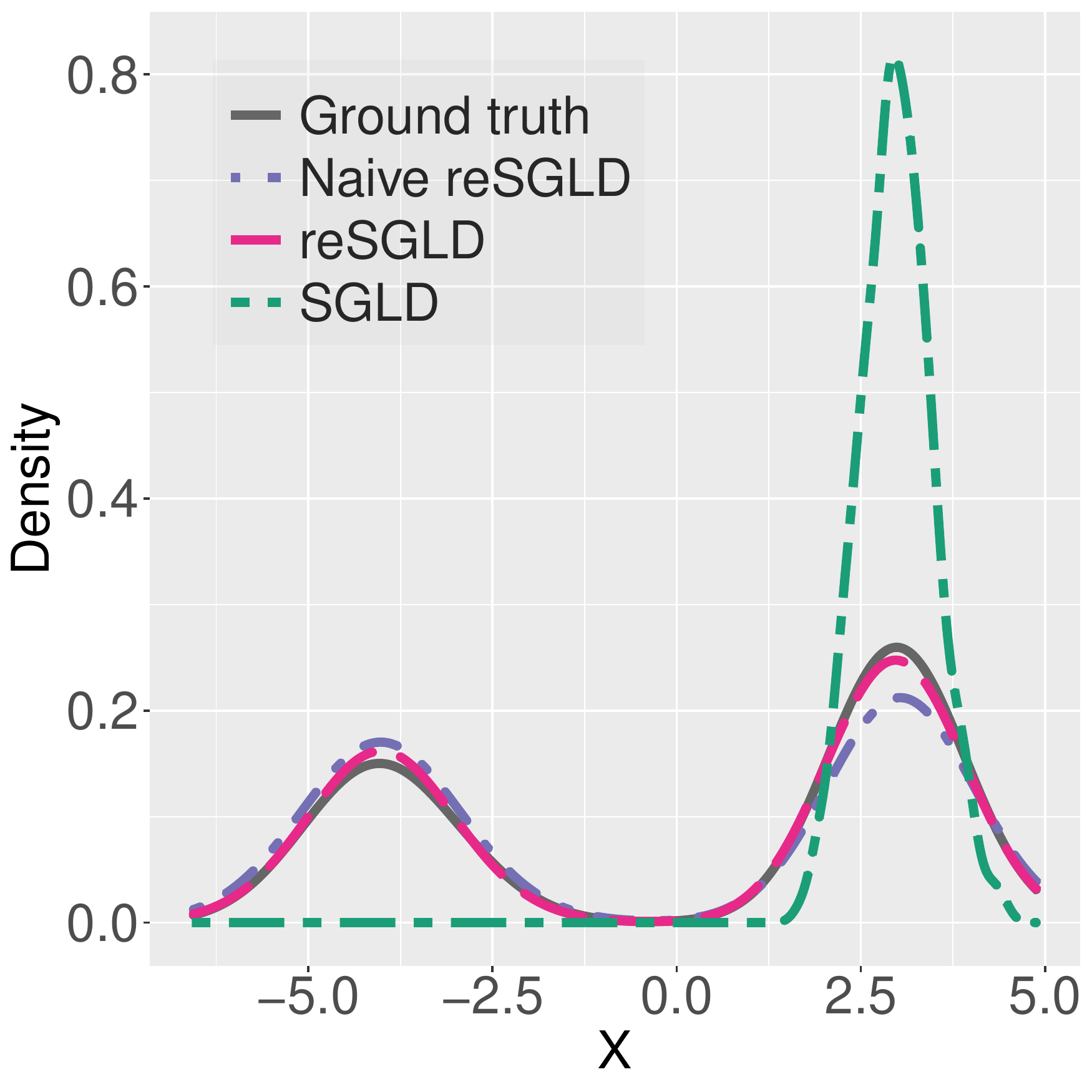}}\label{fig: 22c}\quad
  \subfigure[$\scriptstyle \widetilde U_3(x)\sim U_3(x)+7 t(\nu=10)$.]{\includegraphics[scale=0.22]{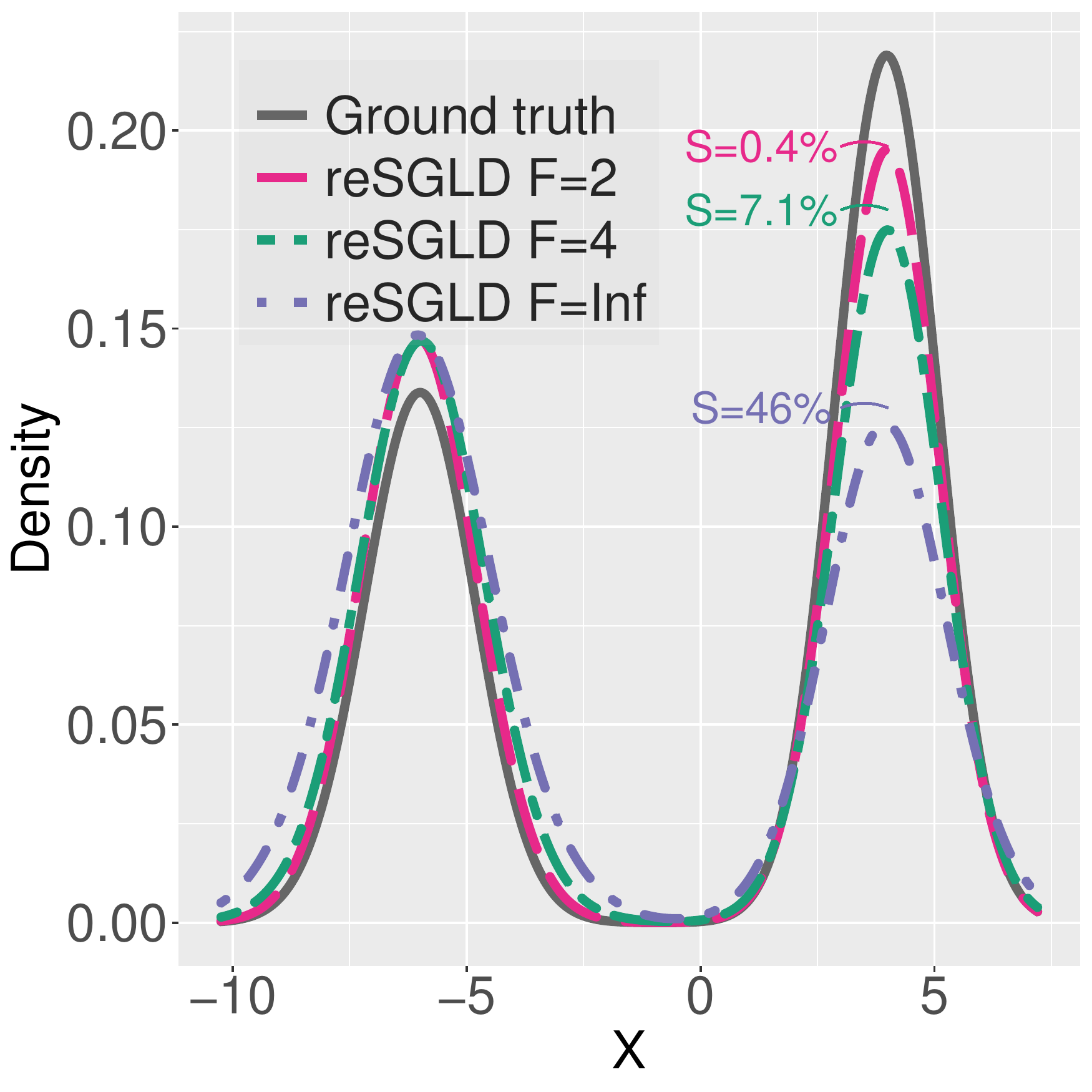}}\label{fig: 22d}
    \vskip -0.15in
  \caption{Evaluation of reSGLD on Gaussian mixture distributions, where the na\"ive reSGLD doesn't make any corrections and reSGLD proposes to adaptively estimate the unknown corrections. In Fig.1(d), we omit SGLD because it gets stuck in a single mode.}\label{mixture}
  \vspace{-1em}
\end{figure*}

We first demonstrate reSGLD on a simple Gaussian mixture distribution $e^{-U_1(x)}\sim 0.4 \mathcal{N}(-3, 0.7^2)+0.6\mathcal{N}(2, 0.5^2)$, where $U_1(x)$ is the energy function. We assume we can only obtain the unbiased energy estimator $\widetilde U_1(x)\sim \mathcal{N}(U_1(x), 2^2)$ and the corresponding stochastic gradient at each iteration.  From Fig.\ref{mixture}(a,b), we see that SGLD suffers from the local trap problem and takes a long time to converge. The na\"ive reSGLD algorithm alleviated the local trap problem, but is still far away from the ground truth distribution without a proper correction. The na\"ive reSGLD converges faster than reSGLD in the early phase due to a higher swapping rate, but ends up with a large bias when the training continues. By contrast, reSGLD successfully identifies the right correction through adaptive estimates and yields a close approximation to the ground truth distribution. The high-temperature chain serves as a bridge to facilitate the movement, and the local trap problem is greatly reduced.

In the second example, we relax the normality assumption to a heavy-tail distribution. Given a Gaussian mixture distribution $e^{-U_2(x)}\sim 0.4 \mathcal{N}(-4, 0.7^2)+0.6\mathcal{N}(3, 0.5^2)$, we assume that we can obtain the stochastic energy estimator $\widetilde U_2(x)\sim U_2(x)+t(\nu=5)$, where $t(\nu=5)$ denotes the Student's t-distribution with degree of freedom 5. We see from Fig.\ref{mixture}(c) that reSGLD still gives a good approximation to the ground true distribution while the others don't.

In the third example, we show a case when the correction factor $F$ is useful. We sample from a Gaussian mixture distribution $e^{-U_3(x)}\sim 0.4 \mathcal{N}(-6, 0.7^2)+0.6\mathcal{N}(4, 0.5^2)$. We note that the two modes are far away from each other and the distribution is more difficult to simulate. More interestingly, we assume $\widetilde U_3(x)\sim U_3(x)+7 t(\nu=10)$, which requires a large correction term and ends up with no swaps in the finite 100,000 iterations at $F=1$. In such a case, the unbiased algorithm behaves like the ordinary SGLD as in Fig.\ref{mixture}(c) and still suffers from the local trap problems. To achieve larger acceleration effects to avoid local traps and maintain accuracy, we try $F$ at $2$, $4$ and $\infty$ ($\text{Inf}$), where the latter is equivalent to the na\"ive reSGLD. We see from Fig.\ref{mixture}(d) that $F=2$ shows the best approximations, despite that the swapping rate $S$ is only $0.4\%$. Further increases on the acceleration effect via larger correction factors $F$ give larger swapping rates (7.1\% and 46\%) and potentially accelerate the convergence in the beginning. However, the biases become more significant as we increase $F$ and lead to larger errors in the end. 

\subsection{Optimization of Image Data}
\label{SL}
We evaluate the adaptive replica exchange Monte Carlo on CIFAR10 and CIFAR100 
, which consist of 50,000 $32\times32$ RGB images for training and 10,000 images for testing. CIFAR10 and CIFAR100 have 10 classes and 100 classes, respectively. We adopt the well-known residual networks (ResNet) 
and wide ResNet (WRN) 
as model architectures. We use the 20, 32, 56-layer
ResNet (denoted as ResNet-20, et al.), WRN-16-8 and WRN-28-10, where, for example, WRN-16-8 denotes a ResNet that has 16 layers and is 8 times wider than the original. Inspired by the popular momentum stochastic gradient descent, we use stochastic gradient Hamiltonian Monte Carlo (SGHMC) as the baseline sampling algorithm and use the numerical method proposed by \citet{Saatci17} to reduce the tuning cost. We refer to the momentum stochastic gradient descent algorithm as M-SGD and the adaptive replica exchange SGHMC algorithm as reSGHMC.

\begin{figure*}[!ht]
  \centering
  \subfigure[Time-varying corrections and losses on CIFAR10]{\includegraphics[width=3.85cm, height=3.3cm]{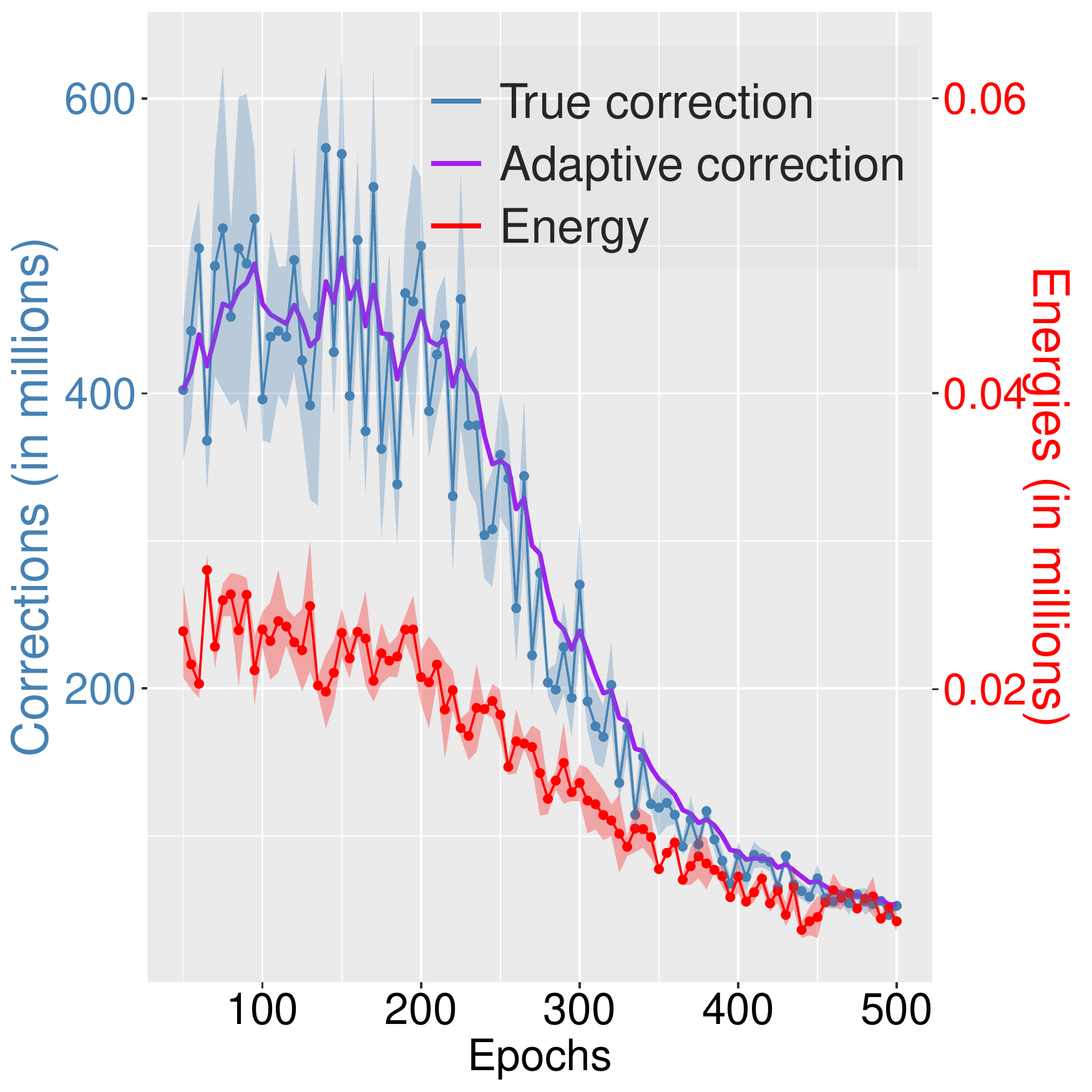}}\label{fig: 2a}\quad
  \subfigure[Time-varying corrections and losses on CIFAR100]{\includegraphics[width=3.85cm, height=3.3cm]{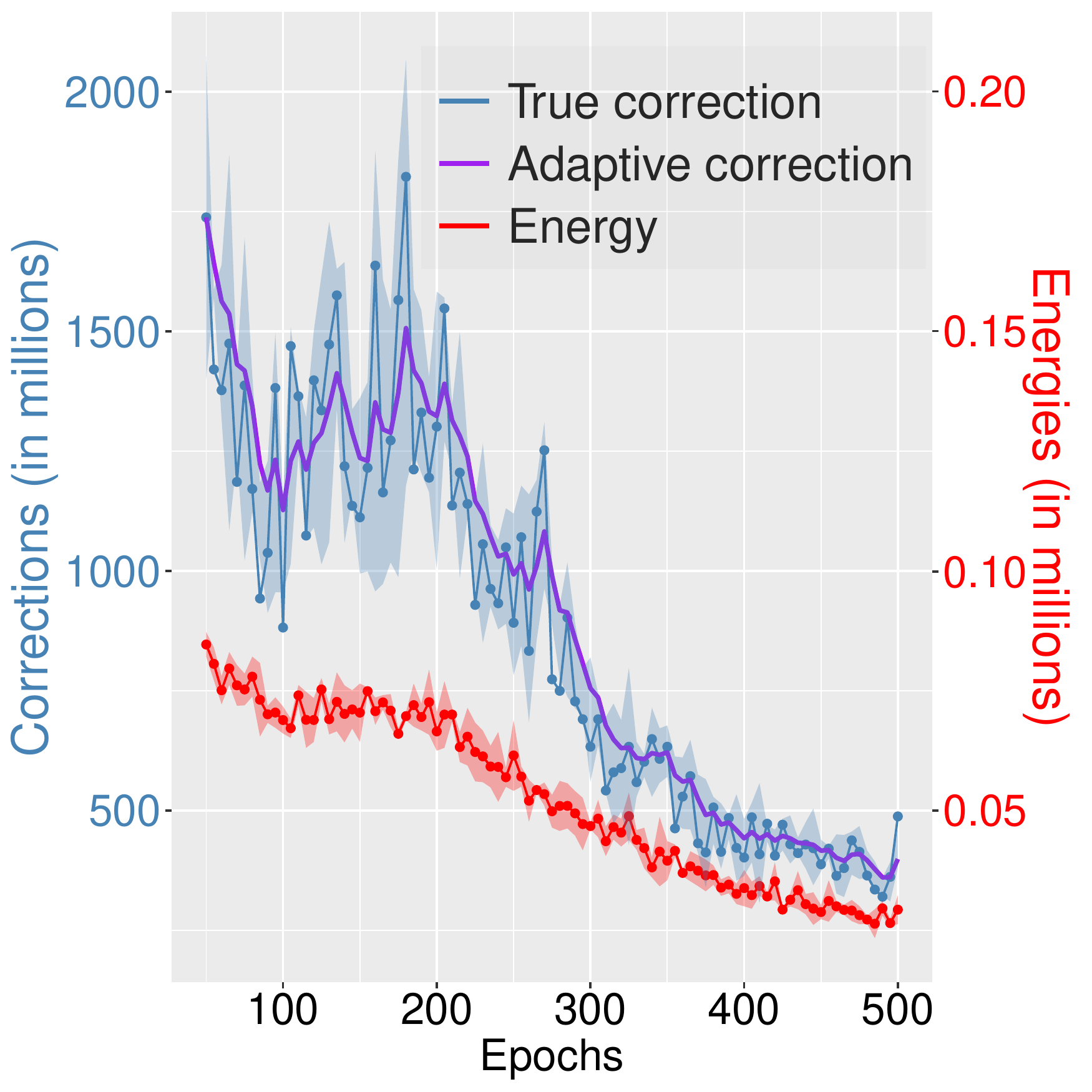}}\label{fig: 2b}\quad
  \subfigure[Effects of correction factors on CIFAR10]{\includegraphics[width=3.5cm, height=3.3cm]{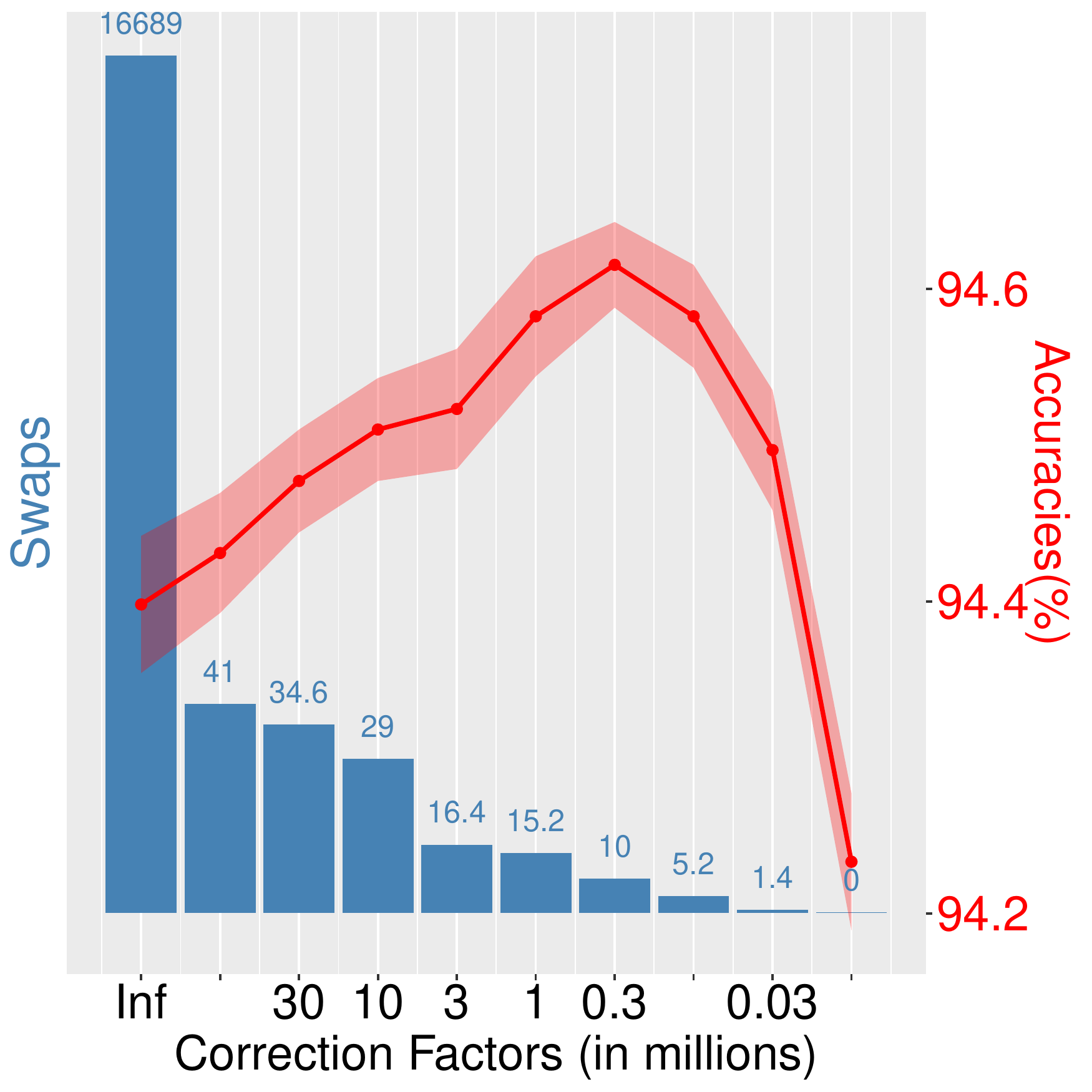}}\label{fig: 2c}\quad
  \subfigure[Effects of correction factors on CIFAR100]{\includegraphics[width=3.5cm, height=3.3cm]{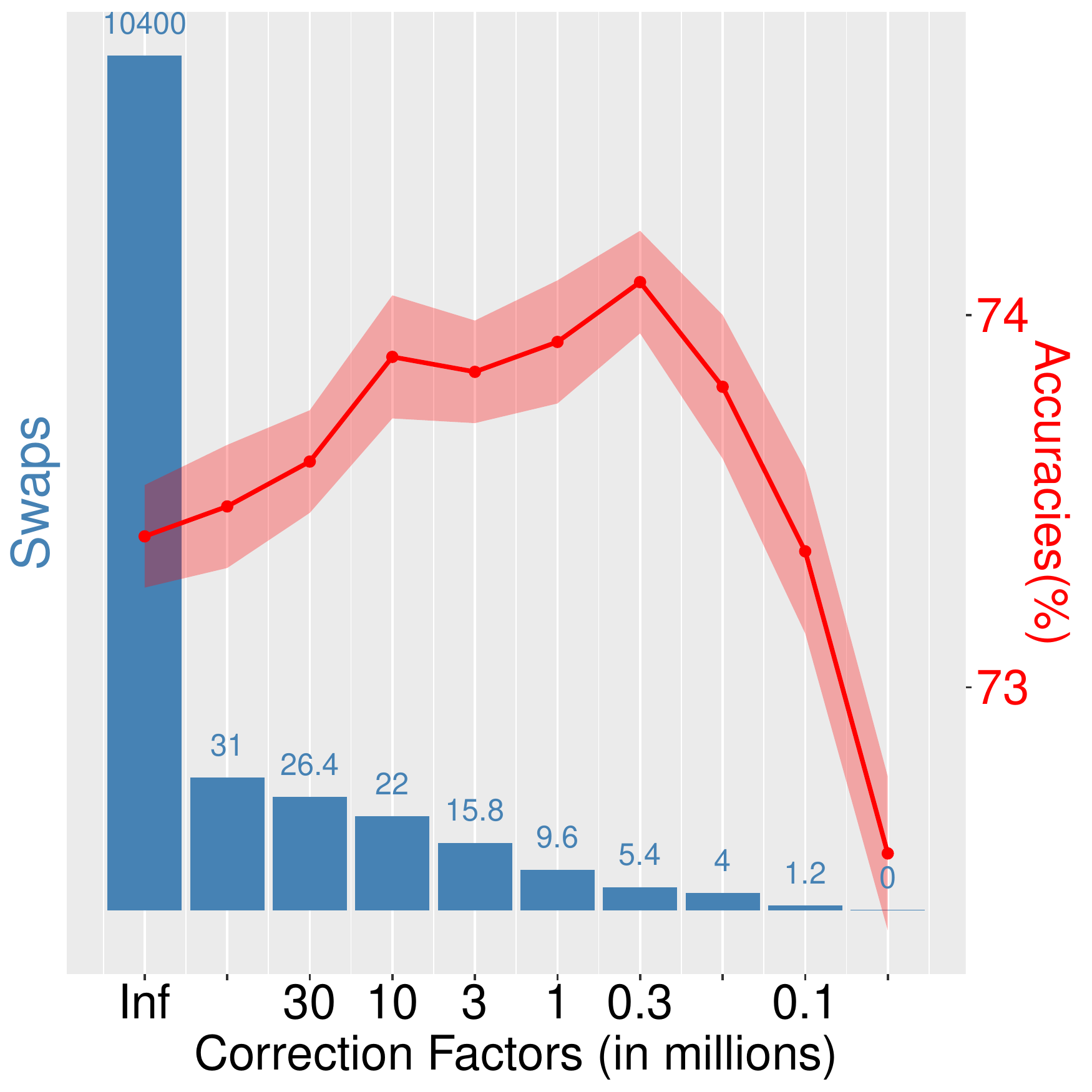}}\label{fig: 2d}
    \vskip -0.15in
  \caption{Time-varying variances of the stochastic energy based on batch-size 256 on CIFAR10 \& CIFAR100 datasets.}
  \label{cifar_biases}
  \vspace{-1.3em}
\end{figure*}

We first run several experiments to study the ideal corrections for the optimization of deep neural networks based on the fixed temperatures $\tau_1=0.01$ and $\tau_2=0.05$. We observe from Fig.(\ref{cifar_biases})(a, b) that the corrections are thousands of times larger than the energy losses, which implies that an exact correction leads to no swaps in practice and no acceleration can be achieved. The desire to obtain more acceleration effects drives us to manually shrink the corrections by increasing $F$ to increase the swapping rates, although we have to suffer from some model bias. 

\begin{table*}[ht]
\begin{sc}
\caption[Table caption text]{Prediction accuracies (\%) with different architectures on CIFAR10 and CIFAR100.}\label{nonconvex_funcs}
\begin{center} 
\begin{tabular}{c|ccc|ccc}
\hline
\multirow{2}{*}{Model} & \multicolumn{3}{c|}{CIFAR10} & \multicolumn{3}{c}{CIFAR100} \\
\cline{2-7}
 & M-SGD & SGHMC & \lowercase{re}{SGHMC} & M-SGD & SGHMC & reSGHMC  \\
\hline
\hline
ResNet-20 & 94.21$\pm$0.16 & 94.22$\pm$0.12 & \textbf{94.62$\pm$0.18} & 72.45$\pm$0.20 & 72.49$\pm$0.18 & \textbf{74.14$\pm$0.22}\\ 
ResNet-32 & 95.15$\pm$0.08 & 95.18$\pm$0.06 & \textbf{95.35$\pm$0.08} & 75.01$\pm$0.22 & 75.14$\pm$0.28 & \textbf{76.55$\pm$0.30} \\ 
ResNet-56 & 96.01$\pm$0.08 & 95.95$\pm$0.10 & \textbf{96.12$\pm$0.06} & 78.96$\pm$0.32 & 79.04$\pm$0.30 & \textbf{80.14$\pm$0.34}    \\
\hline
\hline
WRN-16-8 &  96.71$\pm$0.06 & 96.73$\pm$0.08 & \textbf{96.87$\pm$0.06} & 81.70$\pm$0.26 & 82.07$\pm$0.22 & \textbf{82.95$\pm$0.30}\\
WRN-28-10 & 97.33$\pm$0.08 & 97.32$\pm$0.06 & \textbf{97.42$\pm$0.06} & 83.79$\pm$0.18 & 83.76$\pm$0.14 & \textbf{84.38$\pm$0.18} \\
\hline
\end{tabular}
\end{center} 
\end{sc}
\vspace{-1.5em}
\end{table*}

We study the model performance by applying different correction factors $F$. We choose batch-size 256 and run the experiments within 500 epochs. We first tune the optimal hyperparameters for M-SGD, SGHMC and the low-temperature chain of reSGHMC: we set the learning rate $\eta_k^{(1)}$ to 2e-6 in the first 200 epochs and decay it afterward by a factor of 0.984 every epoch; the low temperature follows an annealing schedule $\tau_1=\frac{0.01}{1.02^k}$ to accelerate the optimization; the weight decay is set to 25. Then, as to the high-temperature chain of reSGHMC, we use a larger learning rate $\eta^{(2)}_k=1.5 \eta^{(1)}_k$ and a higher temperature $\tau_2=5\tau_1$. We set $F$ as $F_0$ in the beginning and then adapt the value to counteract the annealing effect of the temperatures.
The variance estimator is updated each epoch based on the variance of 10 samples of the stochastic energies and the smoothing factor is set to $\gamma=0.3$ in (\ref{smoothing}). Consequently, the computations only increase by less than 5\%. In addition, we use a thinning factor $200$ and report all the results based on Bayesian model averaging. We repeat every experiment five times to obtain the mean and 2 standard deviations.

We see from Fig.\ref{cifar_biases}(c,d) that both datasets rely on a very large initial correction factor $F_0$ to yield good performance and the optimal initial correction factor $\widehat F_0$ is achieved at 3e5. Empirically, we notice that the first five swaps provide the \emph{largest marginal improvement} in acceleration. A larger $F_0$ than $\widehat F_0$ leads to a larger swapping rate with more swaps and thus a larger acceleration effect, however, the performance still decreases as we increase $F_0$, implying that the biases start to dominate the error and the diminishing marginal improvement on the acceleration effect is no longer significant. We note that there is only one extra hyper-parameter, namely, the correction factor $F$, required to tune, and it is independent of the standard SGHMC. This shows that the tuning cost is acceptable. 

\begin{table}[ht]
\begin{sc}
\vskip -0.1in
\caption[Table caption text]{Prediction accuracies (\%) with different batch sizes on CIFAR10 \& CIFAR100 using ResNet-20.}\label{batch_effects}
\begin{center} 
\begin{tabular}{c|ccc}
\hline
Batch & M-SGD & SGHMC & reSGHMC  \\
\hline
\hline
 &\multicolumn{3}{c}{CIFAR10} \\
256 & 94.21$\pm$0.16 & 94.22$\pm$0.12& 94.62$\pm$0.18 \\ 
1024 & 94.49$\pm$0.12 & 94.57$\pm$0.14& \textbf{95.01$\pm$0.16} \\ 
\hline
\hline
 &\multicolumn{3}{c}{CIFAR100} \\
256 & 72.45$\pm$0.20 & 72.49$\pm$0.18 & 74.14$\pm$0.21 \\ 
1024 & 73.31$\pm$0.18 & 73.23$\pm$0.20& \textbf{75.11$\pm$0.26} \\
\hline
\end{tabular}
\end{center} 
\end{sc}
\vspace{-1em}
\end{table}

To obtain a comprehensive evaluation of reSGHMC, we use the optimal correction factor for reSGHMC and test it on ResNet20, 32, 56, WRN-16-8 and WRN-28-10. From Table.\ref{nonconvex_funcs}, we see that reSGHMC consistently outperforms SGHMC and M-SGD on both datasets, showing the robustness of reSGHMC to various model architectures. For CIFAR10, our method works better with ResNet-20 and ResNet-32 and improves the prediction accuracy by 0.4\% and 0.2\%, respectively. Regarding the other model architectures, it still slightly outperforms the baselines by roughly 0.1\%-0.2\%, although this dataset is highly optimized. Specifically, reSGHMC achieves the state-of-the-art 97.42\% accuracy with WRN-28-10 model. For CIFAR100, reSGHMC works particularly well based on various model architectures. It outperforms the baseline by as high as 1.5\% using ResNet-20 and ResNet-32, and around 1\% based on the other architectures. It also achieves the state-of-the-art 84.38\% based on WRN-28-10 on CIFAR100. 

We also conduct the experiments using larger batch sizes with ResNet-20 and report the results in Table.\ref{batch_effects}. We run the same iterations and keep the other setups the same. We find that a larger batch size significantly boosts the performance of reSGHMC by as much as 0.4\% accuracies on CIFAR10 and 1\% on CIFAR100, which shows the \emph{potential of using a large batch size} in the future.

\subsection{Bayesian GANs for Semi-supervised Learning}

\begin{figure*}[!ht]
  \centering
  \subfigure[CIFAR10]{\includegraphics[scale=0.2]{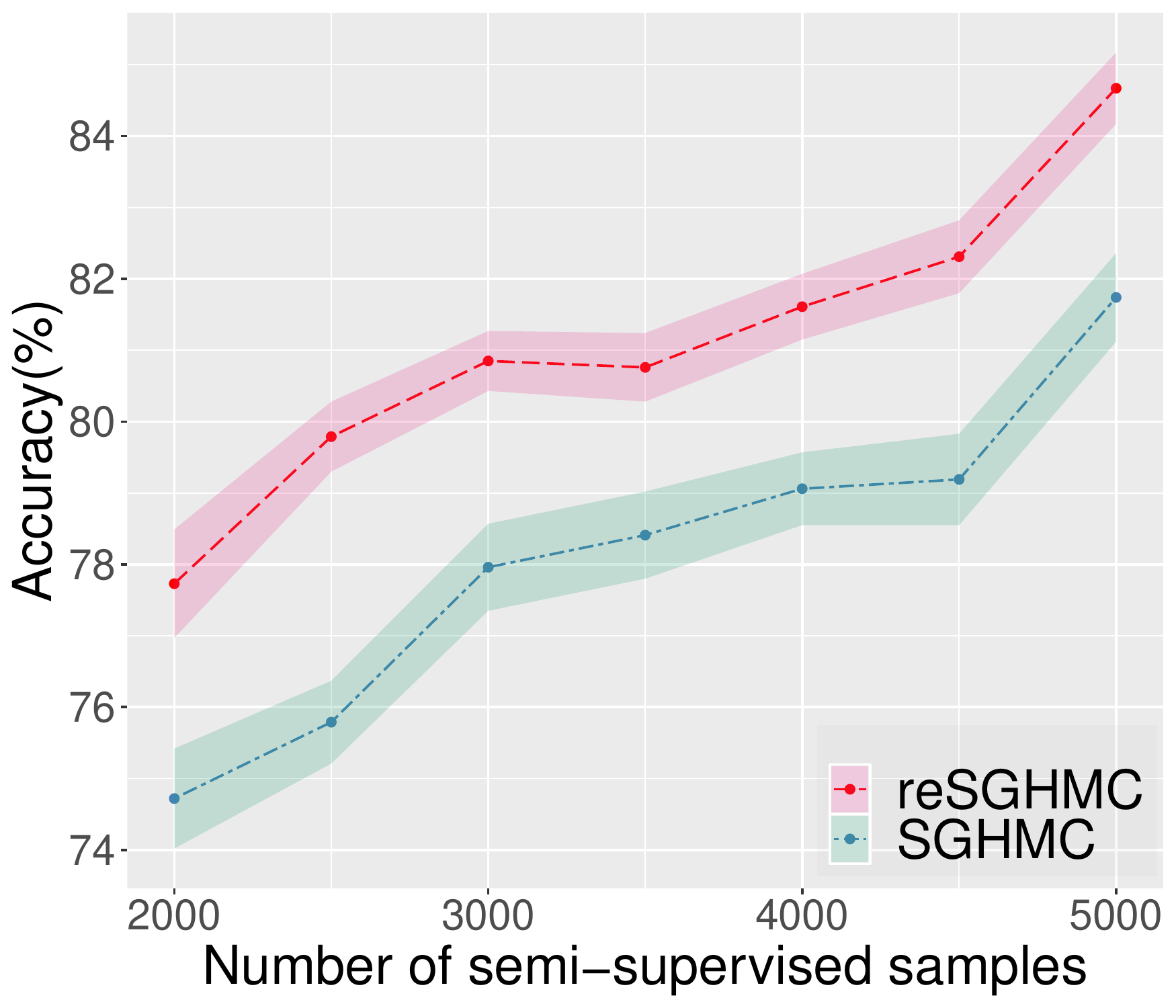}}\label{fig: 3a}\quad\quad
  \hspace{0.3cm}
  \subfigure[CIFAR100]{\includegraphics[scale=0.2]{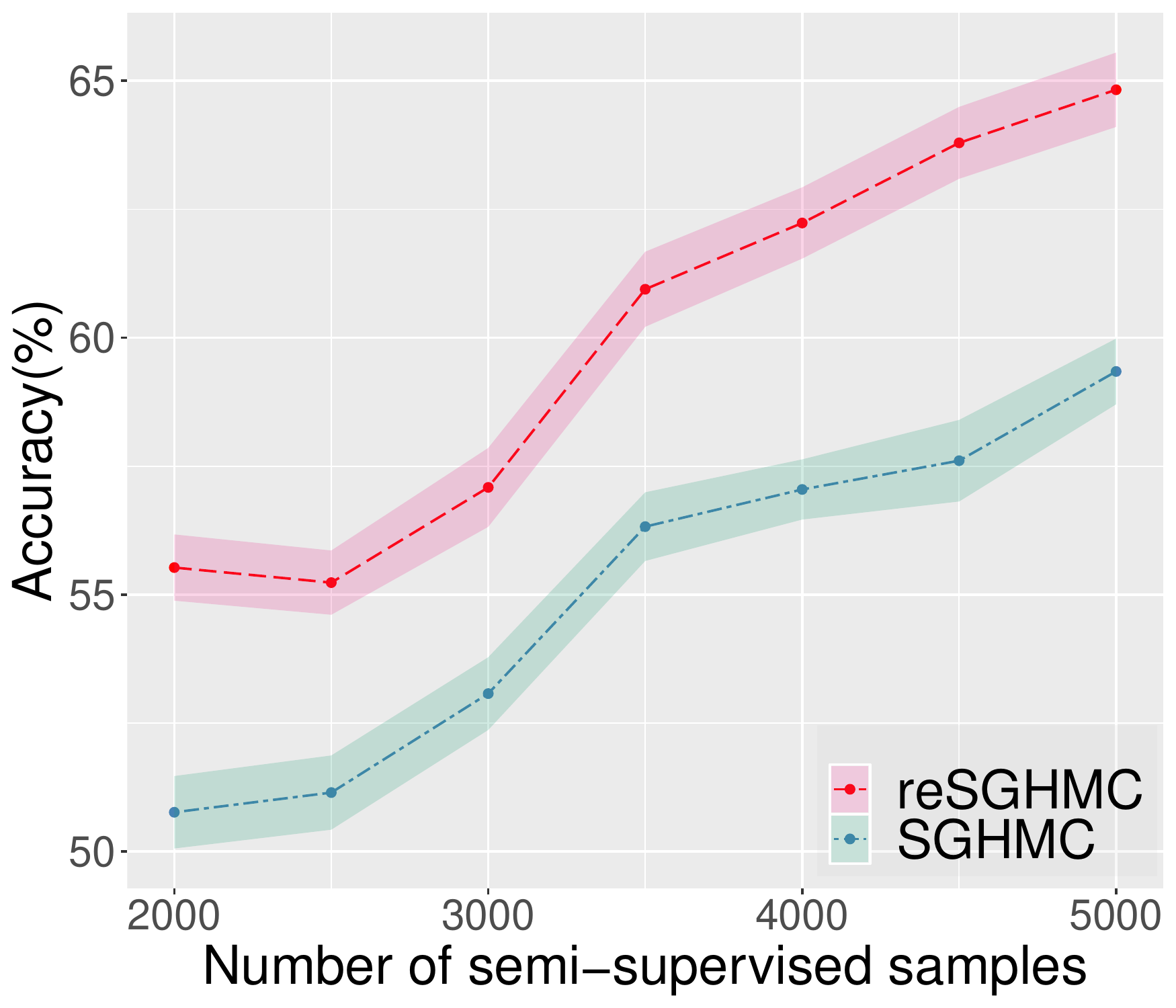}}\label{fig: 3b}\quad\quad
  \hspace{0.3cm}
  \subfigure[SVHN]{\includegraphics[scale=0.2]{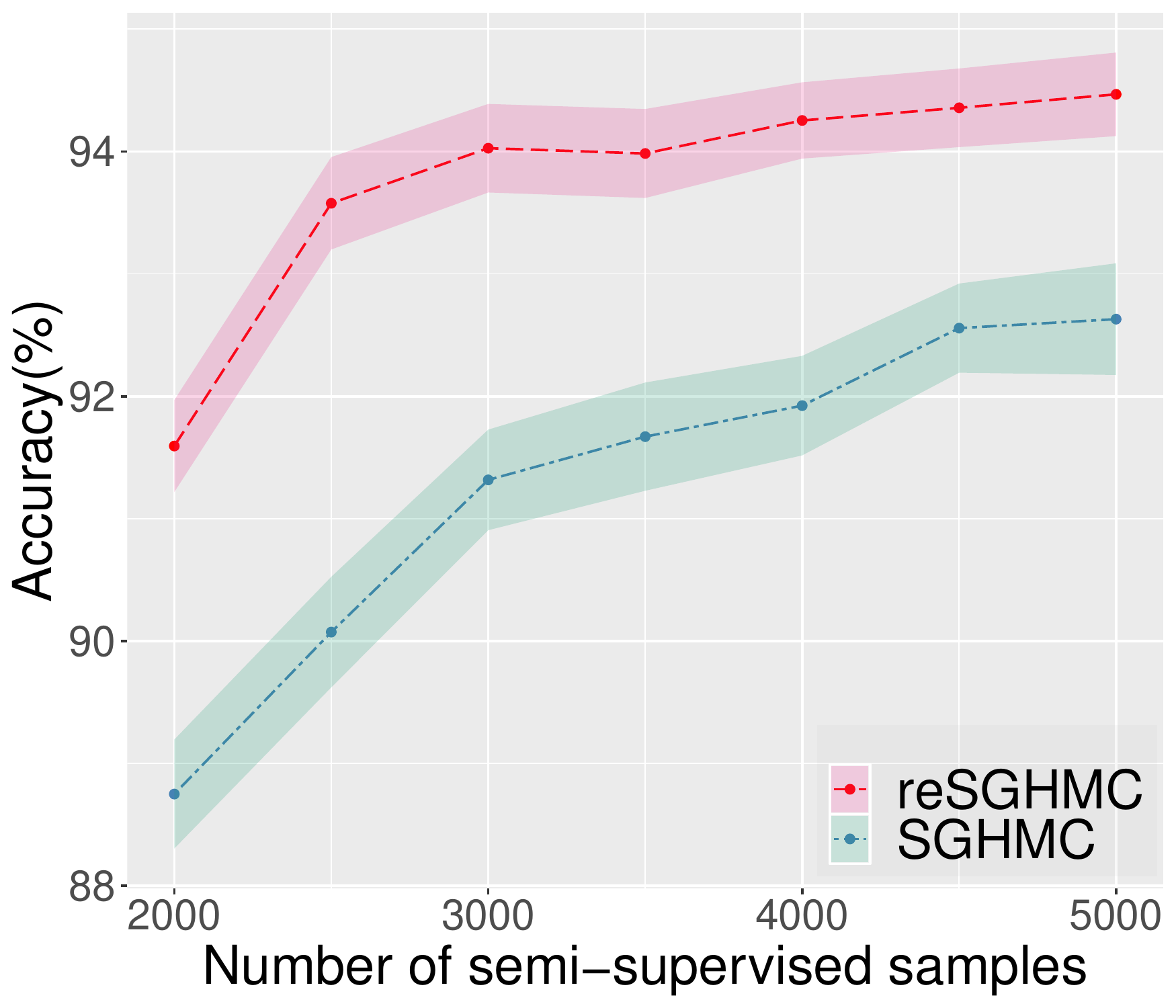}}\label{fig: 3c}
  \vspace{-1em}
  \caption{reSGHMC versus SGHMC on benchmark datasets in semi-supervised learning. }
  \label{OneChainVSTwoChains}
  \vspace{-1em}
\end{figure*}

\begin{table*}[ht]
\begin{sc}
\caption[Table caption text]{Semi-supervised learning on CIFAR10, CIFAR100 and SVHN based on different number of labels.}\label{semi-supervised}
\begin{center} 
\begin{tabular}{c|cc|cc|cc}
\hline
\multirow{2}{*}{$N_s$} & \multicolumn{2}{c|}{CIFAR10} & \multicolumn{2}{c|}{CIFAR100} & \multicolumn{2}{c}{SVHN} \\
\cline{2-7}
 & SGHMC & \lowercase{re}{SGHMC} & SGHMC & \lowercase{re}{SGHMC} & SGHMC & \lowercase{re}{SGHMC}  \\
\hline
\hline
2000 & 74.72$\pm$0.39 & \textbf{77.73$\pm$0.31} & 50.76$\pm$0.71 & \textbf{55.53$\pm$ 0.64} & 88.75$\pm$0.44 & \textbf{91.59$\pm$0.38}\\ 
3000 & 77.96$\pm$0.32 & \textbf{80.85$\pm$0.23} & 53.07$\pm$0.71 & \textbf{57.09$\pm$ 0.77} & 	91.32$\pm$0.41 & \textbf{94.03$\pm$0.36} \\ 
4000 & 79.06$\pm$0.29 & \textbf{81.61$\pm$0.24} & 57.05$\pm$0.59 & \textbf{62.23$\pm$ 0.69} & 91.92$\pm$0.41 & \textbf{94.25$\pm$0.31}    \\
5000 & 81.74$\pm$0.36 & \textbf{84.67$\pm$0.28} & 59.34$\pm$0.64 & \textbf{64.83$\pm$ 0.72} & 92.63$\pm$0.46 & \textbf{94.33$\pm$0.34}\\
\hline
\end{tabular}
\end{center} 
\end{sc}
\vspace{-1.5em}
\end{table*}

Semi-supervised learning (SSL) is an economic machine learning task because it doesn't require all the data to have pricey labels and still shows promising results. However, the multi-modal problem is more severe in the training of SSLs, such as Bayesian GANs \citep{Saatci17}, which motivates us to utilize a more powerful algorithm for multi-modal sampling. Therefore, we further evaluate reSGHMC in SSL on CIFAR10, CIFAR100 and the StreetView House Numbers dataset (SVHN) \footnote[2]{SVHN consists of 73,257 10-class images for training and 26,032 images for testing. } using Bayesian GANs  and study how swaps boost the performance. 

Regarding the Bayesian GANs used for SSLs, we transform the ordinary discriminator into a K+1-class classifier, where $K$ is the number of classes in each dataset, and $K=10$ for CIFAR10 and SVHN and $K=100$ for CIFAR100. During training, a five-layer Bayesian deconvolutional GAN is used as the generator to increase the performance of the discriminator. After training, we discard the generator and use the discriminator for predictions. Since SGHMC is the standard baseline method to simulate from Bayesian GAN, we only compare reSGHMC with SGHMC and no longer report the results based on M-SGD. Following \citet{Saatci17}, we take 10 Monte Carlo (MC) samples for the generator and just 1 MC sample for the discriminator. We also simulate 2 samples from SGHMC or reSGHMC for each MC sample. Moreover, we only use 2 chains in the reSGHMC algorithm.

We study the model performance based on different number of labeled data $N_s$ in semi-supervised learning: $N_s=\{2000, 2500, ..., 5000\}$. Similar to the experiments of supervised learning in Section \ref{SL}, a large correction factor $F$ tends to decrease the performance by injecting more biases, and a smaller correction factor $F$ leads to smaller acceleration effects. Therefore, a good correction factor $F$ is required to balance between the acceleration effects and accuracies. 
We choose batch-size 64 and detail the other hyper-parameter settings in the Appendix. We decay the learning rate while training, but no longer decrease the temperatures. We report each result using Bayesian model average and repeat each experiment five times to get the average and 2 standard deviations. 

As shown in Fig.\ref{OneChainVSTwoChains} and Table.\ref{semi-supervised}, we observe that a larger number of labeled images leads to better performance for all the three datasets. In particular, the 3000 additional labeled images boost the prediction accuracies on CIFAR10, CIFAR100, and SVHN by 7\%, 9\%, and 3\%, respectively. CIFAR10 and CIFAR100 are more sensitive to the labeled images and show larger marginal improvements given a smaller number of labeled images. 

Compared to SGHMC, reSGHMC shows a significantly pronounced difference in performance. The consistent improvements are nearly 3\% for CIFAR10, 5\% for CIFAR100 and 2\% for SVHN, respectively. From Table.\ref{nonconvex_funcs} and Table.\ref{semi-supervised}, the large improvement in SSL indicates that the multi-modal problem is more severe in Bayesian GANs and the high-temperature chain facilitates the low-temperature chain to jump over distinct modes for the exploration of rich multimodal distributions. In the end, the low-temperature chain obtains both the exploration ability to traverse the whole domain and the exploitation ability to explore the local geometry, which greatly avoids the mode collapse problems and enables the state-of-the-art performance in SSL.

%% file: 6.conclusion.tex
We propose the adaptive replica exchange SGMCMC algorithm and prove the accelerated convergence in terms of 2-Wasserstein distance. The theory implies an accuracy-acceleration trade-off and guides us to tune the correction factor $F$ to obtain the optimal performance. We support our theory with extensive experiments and obtain significant improvements over the vanilla SGMCMC algorithms on CIFAR10, CIFAR100, and SVHN.

For future works, it is promising to relax the asymptotic normality assumption to the heavy-tailed generalization of L\'{e}vy-stable distribution \citep{Simsekli2019b} and apply It\^{o}'s lemma to geometric L\'{e}vy process to analyze the bias from fat-tailed noises with small batch sizes. Besides, variance reduction \cite{Xu18} of the stochastic noise to obtain a larger acceleration effect is also appealing in both theory and practice. From the computational perspective, it is also interesting to study parallel multi-chain reSGMCMC in larger machine learning tasks. 

%% file: 7.acknowledge.tex
We thank the reviewers for their suggestions. We acknowledge the support from the Bilsland Dissertation Fellowship (Deng), the National Science Foundation DMS-1555072, DMS-1736364, DMS-1821233 (Lin) and DMS-1818674 (Liang) and the GPU grant program from NVIDIA.

%% file: supp.tex

\newcommand{\la}{\langle}
\newcommand{\ra}{\rangle}
\newcommand{\hP}{\hat\dbP}
\newcommand{\vr}{\overrightarrow}
\newcommand{\wt}{\widetilde}
\newcommand{\dd}{\mathcal{\dagger}}
\newcommand{\ts}{\mathsf{T}}
\newcommand{\tone}{t_{1}}
\newcommand{\ttwo}{t_{2}}
\newcommand{\tthree}{t_{3}}
\newcommand{\tfour}{t_{4}}
\newcommand{\pn}{P^{\tau}\nabla}


\newcommand{\oo}{\overline}
\newcommand{\eps}{\varepsilon}
\newcommand{\LDP}{{{LDP }}}
\newcommand{\MDP}{{{MDP }}}
\newcommand{\LDPx}{{{LDP}}}
\newcommand{\Las}{{\La^*}}
\newcommand{\ba}{\begin{array}}
\newcommand{\ea}{\end{array}}
\newcommand{\be}{\begin{equation}}
\newcommand{\ee}{\end{equation}}
\newcommand{\bea}{\begin{eqnarray}}
\newcommand{\eea}{\end{eqnarray}}
\newcommand{\beaa}{\begin{eqnarray*}}
\newcommand{\eeaa}{\end{eqnarray*}}
\newcommand{\Remark}{\noindent{\bf Remark:}\  }
\newcommand{\Remarks}{\noindent{\bf Remarks:}\  }
\newcommand{\iL}{(L(\rho))^{-1}}

\def\dbB{\mathbb{B}}
\def\dbC{\mathbb{C}}
\def\dbD{\mathbb{D}}
\def\dbE{\mathbb{E}}
\def\dbF{\mathbb{F}}
\def\dbG{\mathbb{G}}
\def\dbH{\mathbb{H}}
\def\dbI{\mathbb{I}}
\def\dbJ{\mathbb{J}}
\def\dbK{\mathbb{K}}
\def\dbL{\mathbb{L}}
\def\dbM{\mathbb{M}}
\def\dbN{\mathbb{N}}
\def\dbP{\mathbb{P}}
\def\dbR{\mathbb{R}}
\def\dbS{\mathbb{S}}
\def\dbQ{\mathbb{Q}}

\def\Dom{{\rm dom}}

\def\a{\alpha}
\def\b{\beta}
\def\g{\gamma}
\def\d{\delta}
\def\e{\varepsilon}
\def\z{\zeta}
\def\k{\kappa}
\def\l{\lambda}
\def\m{\mu}
\def\n{\nu}
\def\si{\sigma}
\def\t{\tau}
\def\f{\varphi}
\def\th{\theta}
\def\h{\widehat}
%
%
%
\def\G{\Gamma}
\def\D{\Delta}
\def\Th{\Theta}
\def\L{\Lambda}
\def\Si{\Sigma}
\def\F{\Phi}
%
%
\def\cA{{\cal A}}
\def\cB{{\cal B}}
\def\cC{{\cal C}}
\def\cD{{\cal D}}
\def\cE{{\cal E}}
\def\cF{{\cal F}}
\def\cG{{\cal G}}
\def\cH{{\cal H}}
\def\cI{{\cal I}}
\def\cJ{{\cal J}}
\def\cK{{\cal K}}
\def\cL{{\cal L}}
\def\cM{{\cal M}}
\def\cN{{\cal N}}
\def\cO{{\cal O}}
\def\cP{{\cal P}}
\def\cQ{{\cal Q}}
\def\cR{{\cal R}}
\def\cS{{\cal S}}
\def\cT{{\cal T}}
\def\cU{{\cal U}}
\def\cV{{\cal V}}
\def\cW{{\cal W}}
\def\cX{{\cal X}}
\def\cY{{\cal Y}}
\def\cZ{{\cal Z}}
%
\def\hA{\mathbb{A}}
\def\hB{\mathbb{B}}
\def\hC{\mathbb{C}}
\def\hD{\mathbb{D}}
\def\hE{\mathbb{E}}
\def\hF{\mathbb{F}}
\def\hG{\mathbb{G}}
\def\hH{\mathbb{H}}
\def\hI{\mathbb{I}}
\def\hJ{\mathbb{J}}
\def\hK{\mathbb{K}}
\def\hL{\mathbb{L}}
\def\hM{\mathbb{M}}
\def\hN{\mathbb{N}}
\def\hO{\mathbb{O}}
\def\hP{\mathbb{P}}
\def\hQ{\mathbb{Q}}
\def\hR{\mathbb{R}}
\def\hS{\mathbb{S}}
\def\hT{\mathbb{T}}
\def\hU{\mathbb{U}}
\def\hV{\mathbb{V}}
\def\hW{\mathbb{W}}
\def\hX{\mathbb{X}}
\def\hY{\mathbb{Y}}
\def\hZ{\mathbb{Z}}

\def\sA{\mathscr{A}}
\def\sB{\mathscr{B}}
\def\sC{\mathscr{C}}
\def\sD{\mathscr{D}}
\def\sE{\mathscr{E}}
\def\scF{\mathscr{F}}
\def\sG{\mathscr{G}}
\def\sH{\mathscr{H}}
\def\sI{\mathscr{I}}
\def\sJ{\mathscr{J}}
\def\sK{\mathscr{K}}
\def\scL{\mathscr{L}}
\def\sM{\mathscr{M}}
\def\sN{\mathscr{N}}
\def\sO{\mathscr{O}}
\def\sP{\mathscr{P}}
\def\sQ{\mathscr{Q}}
\def\sR{\mathscr{R}}
\def\sS{\mathscr{S}}
\def\sT{\mathscr{T}}
\def\sU{\mathscr{U}}
\def\sV{\mathscr{V}}
\def\sW{\mathscr{W}}
\def\sX{\mathscr{X}}
\def\sY{\mathscr{Y}}
\def\sZ{\mathscr{Z}}
\def\no{\noindent}
\def\eq{\eqalign}
\def\ss{\smallskip}
\def\ms{\medskip}
\def\bs{\bigskip}
\def\q{\quad}
\def\qq{\qquad}
\def\hb{\hbox}
\def\pa{\partial}
\def\cd{\cdot}
\def\cds{\cdots}
\def\lan{{\langle}}
\def\ran{{\rangle}}
\def\td{\nabla}
\def\bD{{\bf D}}
\def\bF{{\bf F}}
\def\bG{{\bf G}}
\def\bx{{\bf x}}
\def\by{{\bf y}}
\def\bz{{\bf z}}

\def\tr{\hbox{\rm tr}}

\newcommand{\dfnn}{\stackrel{\triangle}{=}}
\newcommand{\basa}{\begin{assumption}}
\newcommand{\easa}{\end{assumption}}
\newcommand{\tbar}{\overline{t}}
\newcommand{\xbar}{\overline{x}}
\newcommand{\bas}{\begin{assum}}
\newcommand{\eas}{\end{assum}}
\newcommand{\lime}{\lim_{\epsilon \rightarrow 0}}
\newcommand{\zep}{z^\epsilon}
\newcommand{\bep}{b^\epsilon}
\newcommand{\hbep}{\hat{b}^\epsilon}
\newcommand{\half}{\frac{1}{2}}

\def\limsup{\mathop{\overline{\rm lim}}}
\def\liminf{\mathop{\underline{\rm lim}}}
\def\ua{\mathop{\uparrow}}
\def\da{\mathop{\downarrow}}
\def\Ra{\mathop{\Rightarrow}}
\def\La{\mathop{\Leftarrow}}
\def\lan{\mathop{\langle}}
\def\ran{\mathop{\rangle}}
\def\embed{\mathop{\hookrightarrow}}
\def\esssup{\mathop{\rm esssup}}
\def\essinf{\mathop{\rm essinf}}
\def\limw{\mathop{\buildrel w\over\rightharpoonup}}
\def\limws{\mathop{\buildrel *\over\rightharpoonup}}
\def\lims{\mathop{\buildrel s\over\rightarrow}} 
\def\limP2{\,\mathop{\buildrel \Pi_2\over\longrightarrow\,}}
\def\lq{\leqno}
\def\rq{\eqno}
\def\pa{\partial}
\def\h{\widehat}
\def\wt{\widetilde}
\def\cds{\cdots}
\def\ae{\hbox{\rm -a.e.{ }}}
\def\as{\hbox{\rm -a.s.{ }}}
\def\sgn{\hbox{\rm sgn$\,$}}
\def\meas{\hbox{\rm meas$\,$}}
\def\supp{\hbox{\rm supp$\,$}}
\def\co{\mathop{{\rm co}}}
\def\coh{\mathop{\overline{\rm co}}}
\def\cl{\overline}
\def\codim{\hbox{\rm codim$\,$}}
\def\Int{\hbox{\rm Int$\,$}}
\def\diam{\hbox{\rm diam$\,$}}
\def\deq{\mathop{\buildrel\D\over=}}
\def\tr{\hbox{\rm tr$\,$}}
\def\deq{\mathop{\buildrel\D\over=}}
\def\Re{\hbox{\rm Re$\,$}}
\def\bnm{{\,|\neg\neg|\neg\neg|\neg\neg|\,}}

\def\ind{{\perp\neg\neg\neg\perp}}

\def\dis{\displaystyle}
\def\wt{\widetilde}
\def\dh{\dot{h}}
\def\dF{\dot{F}}
\def\bF{{\bf F}}
\def\bx{{\bf x}}
\def\cad{c\`{a}dl\`{a}g}
\def\cag{c\`{a}gl\`{a}d~}
\def\bP{{\bf P}}
\def\1{{\bf 1}}
\def\by{{\bf y}}

\def\hSM{\widehat {\cS\!\cM}^2}

\def\:{\!:\!}
\def\reff#1{{\rm(\ref{#1})}}
\def \proof{{\noindent \bf Proof\quad}}

\begin{Large}
\begin{center}
    \textbf{Supplimentary Material for \textit{``Non-convex Learning via Replica Exchange Stochastic Gradient MCMC''}}
\end{center}
\end{Large}

In this supplementary material, we prove the convergence in $\S$\ref{converge} and show the experimental setup in $\S$\ref{setup}.

\section{Convergence Analysis}
\label{converge}
\input{supp_A}

$\newline$
$\newline$
$\newline$
$\newline$
\section{Hyper-parameter Setting for Bayesian GANs}
\label{setup}
\input{supp_B}

%% file: supp_A.tex
\subsection{Background}
The continuous-time replica exchange Langevin diffusion (reLD) $\{ \bbeta_t\}_{t\ge 0 }:=\left\{\begin{pmatrix}{}
\bbeta_t^{(1)}\\
\bbeta_t^{(2)}
\end{pmatrix}\right\}_{t\ge 0 }$ is a Markov process compounded with a Poisson jump process. In particular, the Markov process follows the stochastic differential equations
\begin{equation}
\label{sde_2}
\begin{split}
    d \bbeta^{(1)}_t &= - \nabla U(\bbeta_t^{(1)}) dt+\sqrt{2\tau_1} d\bW_t^{(1)}\\
    d \bbeta^{(2)}_t &= - \nabla U(\bbeta_t^{(2)}) dt+\sqrt{2\tau_2} d\bW_t^{(2)},\\
\end{split}
\end{equation}
where $\bbeta_t^{(1)}, \bbeta_t^{(2)}$ are the particles (parameters) at time $t$ in $\mathbb{R}^d$, $\bW^{(1)}, \bW^{(2)}\in\mathbb{R}^d$ are two independent Brownian motions, $U:\mathbb{R}^d\rightarrow \mathbb{R}$ is the energy function, $\tau_1<\tau_2$ are the temperatures. The jumps originate from the swaps of particles $\bbeta_t^{(1)}$ and $\bbeta_t^{(2)}$ and follow a Poisson process where the jump rate is specified as the Metropolis form $r (1\wedge S(\bbeta_t^{(1)}, \bbeta_t^{(2)}))dt$. Here $r\geq 0$ is a constant, and $S$ follows
\begin{equation*}
\label{swap_1}
\begin{split}
    S(\bbeta_t^{(1)}, \bbeta_t^{(2)})=e^{ \left(\frac{1}{\tau_1}-\frac{1}{\tau_2}\right)\left(U(\bbeta_t^{(1)})-U(\bbeta_t^{(2)})\right)}.\\
\end{split}
\end{equation*}
Under such a swapping rate, the probability $\nu_t$ associated with reLD at time $t$ is known to converge to the invariant measure (Gibbs distribution) with density
\begin{equation*}
\label{pt_density}
\begin{split}
    \pi(\bbeta^{(1)}, \bbeta^{(2)})\propto e^{-\frac{U(\bbeta^{(1)})}{\tau_1}-\frac{U(\bbeta^{(2)})}{\tau_2}}.
\end{split}
\end{equation*}

In practice, obtaining the exact energy and gradient for reLD (\ref{sde_2}) in a large dataset is quite expensive. We consider the replica exchange stochastic gradient Langevin dynamics (reSGLD), which generates iterates $\{\widetilde \bbeta^{\eta}(k)\}_{k\ge 1}$ as follows
\begin{equation}
\label{reSGLD}
       \begin{split}
           \widetilde \bbeta^{\eta(1)}(k+1) &= \widetilde \bbeta^{\eta(1)}(k)- \eta \nabla \widetilde U (\widetilde \bbeta^{\eta(1)}(k))+\sqrt{2\eta\tau_1} \bxi_k^{(1)}\\
    \widetilde \bbeta^{\eta(2)}(k+1) &= \widetilde \bbeta^{\eta(2)}(k) - \eta\nabla \widetilde U(\widetilde \bbeta^{\eta(2)}(k))+\sqrt{2\eta\tau_2} \bxi_k^{(2)},\\
       \end{split}
   \end{equation}
where $\eta$ is considered to be a fixed learning rate for ease of analysis, and $\bxi_k^{(1)}$ and $\bxi_k^{(2)}$ are independent Gaussian random vectors in $\mathbb{R}^d$. Moreover, the positions of the particles swap based on the stochastic swapping rate. In particular, $\widetilde S(\bbeta^{(1)}, \bbeta^{(2)}):=S(\bbeta^{(1)}, \bbeta^{(2)})+\psi$, and the stochastic gradient $\nabla\widetilde U(\cdot)$ can be written as $\nabla U(\cdot)+\bphi$, where both $\psi\in \mathbb{R}^1$ and $\bphi\in\mathbb{R}^d$ are random variables with mean not necessarily zero. We also denote $\m_k$ as the probability measure associated with $\{\widetilde \bbeta^{\eta}(k)\}_{k\ge 1}$ in reSGLD (\ref{reSGLD}) at step $k$, which is close to $\nu_{k\eta}$ in a suitable sense.

\subsection{Overview of the analysis}

We aim to study the convergence analysis of the probability measure $\m_{k}$ to the invariant measure $\pi$ in terms of 2-Wasserstein distance,
\begin{equation}
\label{w2}
    \mathcal{W}_2(\m, \nu):=\inf_{\Gamma\in \text{Couplings}(\m, \n)}{\sqrt{\int\|\bbeta_{\m}-\bbeta_{\n}\|^2 d \Gamma(\bbeta_{\m},\bbeta_{\n})}},
\end{equation}
where $\|\cdot\|$ is the Euclidean norm, and the infimum is taken over all joint distributions $\Gamma(\bbeta_{\mu}, \bbeta_{\nu})$ with $\mu$ and $\nu$ being the marginals distributions.

By the triangle inequality, we easily obtain that for any $k\in \mathbb{N}$ and $t=k\eta$, we have 
\begin{equation*}
    \mathcal{W}_2(\m_k, \pi)\le \underbrace{\mathcal{W}_2(\m_k, \nu_{t})}_{\text{Discretization error}} +\underbrace{\mathcal{W}_2(\nu_{t}, \pi)}_{\text{Exponential decay}}.
\end{equation*}
We start with the discretization error first by analyzing how reSGLD (\ref{reSGLD}) tracks the reLD (\ref{sde_2}) in 2-Wasserstein distance. The critical part is to study the discretization of the Poisson jump process in mini-batch settings. To handle this issue, we follow \citet{Paul12} and view the swaps of positions as swaps of temperatures. Then we apply standard techniques in stochastic calculus \citep{chen2018accelerating, yin_zhu_10, Issei14, Maxim17} to discretize the Langevin diffusion and derive the corresponding discretization error.

Next, we quantify the evolution of the 2-Wasserstein distance between $\n_t$ and $\pi$. The key tool is the exponential decay of entropy (Kullback-Leibler divergence) when $\pi$ satisfies the log-Sobolev inequality (LSI) \citep{Bakry2014}. To justify LSI, we first verify LSI for reSGLD without swaps, which is a direct result given a proper Lyapunov function criterion \citep{Cattiaux2010} and the Poincar\'{e} inequality \citep{chen2018accelerating}. Then we follow \citet{chen2018accelerating} and verify LSI for reLD with swaps by analyzing the Dirichlet form. Finally, the exponential decay of the 2-Wasserstein distance follows from the Otto-Villani theorem by connecting the 2-Wasserstein distance with the entropy \citep{Bakry2014}.

Before we move forward, we first lay out the following assumptions:
\begin{assumption}[Smoothness]\label{assump: lip and alpha beta}
The energy function $U(\cdot)$ is $C$-smoothness, which implies that there exists a Lipschitz constant $C>0$, such that for every $x,y\in\hR^d$, we have $\|\nabla U(x)-\nabla U(y)\|\le C\|x-y\|.$ \footnote{$\|\cdot\|$ denotes the Euclidean $L^2$ norm.}
\end{assumption}{}

\begin{assumption}[Dissipativity]\label{assump: dissipitive}
The energy function $U(\cdot)$ is $(a,b)$-dissipative, i.e. there exist constants $a>0$ and $b\ge 0$ such that $\forall x\in\mathbb R^d$,  $\la x,\nabla U(x)\ra \ge a\|x\|^2-b.$
\end{assumption}{}

Here the smoothness assumption is quite standard in studying the convergence of SGLD, and the dissipativity condition is widely used in proving the geometric ergodicity of dynamic systems \citep{Maxim17, Xu18}. Moreover, the convexity assumption is not required in our theory.


\subsection{Analysis of discretization error}

The key to deriving the discretization error is to view the swaps of positions as swaps of the temperatures, which has been proven equivalent in distribution \citep{Paul12}. Therefore, we model reLD using the following SDE, 
\begin{equation}\label{replica exchange}
    d\bbeta_t=-\nabla G(\bbeta_t)dt+\Si_td\bW_t,
\end{equation}{}
where  $G(\bbeta_t)=\begin{pmatrix}{}
U(\bbeta_t^{(1)})\\
U(\bbeta_t^{(1)})
\end{pmatrix}$, $\bW\in\mathbb{R}^{2d}$ is a Brownian motion, $\Si_t$ is a random matrix in continuous-time that swaps between the diagonal matrices $\mathbb{M}_1=\begin{pmatrix}{}
\sqrt{2\tau_1}\mathbf I_d&0\\
0&\sqrt{2\tau_2}\mathbf I_d
\end{pmatrix}$ and $\mathbb{M}_2=\begin{pmatrix}{}
\sqrt{2\tau_2}\mathbf I_d&0\\
0&\sqrt{2\tau_1}\mathbf I_d
\end{pmatrix}$ with probability $r S(\bbeta_t^{(1)}, \bbeta_t^{(2)})dt$, and $\mathbf I_d\in \mathbb R^{d\times d}$ is denoted as the identity matrix.

Moreover, the corresponding discretization of replica exchange SGLD (reSGLD) follows:
\begin{equation}
\begin{split}
\label{resgld_2}
    \widetilde \bbeta^{\eta}(k+1)=\widetilde \bbeta^{\eta}(k)-
    \eta\nabla \widetilde G(\widetilde \bbeta^{\eta}(k)) + \sqrt{\eta}\widetilde \Si^{\eta}(k)\bxi_k,
\end{split}
\end{equation}
where $\bxi_k$ is a standard Gaussian distribution in $\mathbb{R}^{2d}$, and $\widetilde \Si^{\eta}(k)$ is a random matrix in discrete-time that swaps between $\mathbb{M}_1$ and $\mathbb{M}_2$ with probability $r\widetilde S(\widetilde \bbeta^{\eta(1)}(k), \widetilde \bbeta^{\eta(2)}(k))\eta$. We denote $\{\widetilde \bbeta_t^{\eta}\}_{t\ge 0 }$ as the continuous-time interpolation of $\{\widetilde \bbeta^{\eta}(k)\}_{k\ge 1}$, which satisfies the following SDE, 
\begin{equation}\label{SGD continuous time interpolation}
	\widetilde \bbeta_t^{\eta}=\widetilde \bbeta_0-\int_0^t\nabla \widetilde G(\widetilde \bbeta^{\eta}_{\lfloor s/\eta \rfloor \eta})ds+\int_0^t\widetilde\Si^{\eta}_{\lfloor s/\eta \rfloor\eta}d\bW_s.
\end{equation}
Here the random matrix $\widetilde \Si_{\lfloor s/\eta\rfloor \eta}^{\eta}$ follows a similar trajectory as $\widetilde \Si^{\eta}(\lfloor s/\eta \rfloor)$. 
For $k\in \mathbb N^{+}$ with $t=k\eta$, the relation $\widetilde \bbeta_t^{\eta}=\widetilde \bbeta_{k\eta}^{\eta}=\widetilde \bbeta^{\eta}(k)$ follows.

\begin{lemma}[Discretization error]\label{discretization}
Given the smoothness and dissipativity assumptions \eqref{assump: lip and alpha beta} and \eqref{assump: dissipitive}, and the learning rate $\eta$ satisfying $0<\eta<1 \land a/C^2$, there exists constants $D_1, D_2$ and $D_3$ such that
\begin{equation}
	\begin{split}
&	\hE[\sup_{0\le t\le T}\|\bbeta_t-\widetilde \bbeta^{\eta}_t||^2] \le D_1 \eta + D_2 \max_{k}\hE[\|\bphi_k\|^2]+D_3\max_{k}\sqrt{\hE\left[|\psi_{k}|^2\right]},\\
	\end{split}
\end{equation}
where $ D_1$ depends on $\tau_1,\tau_2,d, T, C,a,b$; $D_2$ depends on $T$ and $C$; $D_3$ depends on $r, d, T$ and $C$.
\end{lemma}{}

\begin{proof}
Based on the replica exchange Langevin diffusion $\{\bbeta_t\}_{t\ge 0}$ and the continuous-time interpolation of the stochastic gradient Langevin diffusion $\{\widetilde \bbeta_t^{\eta}\}_{t\ge 0}$, we have the following SDE for the difference $\bbeta_t-\widetilde \bbeta^{\eta}_t$. For any $t\in [0,T]$, we have
\begin{equation*}
	\begin{split}
		\bbeta_t-\widetilde \bbeta^{\eta}_t&=-\int_0^t(\nabla G(\bbeta_s)-\nabla\widetilde G(\widetilde \bbeta^{\eta}_{\lfloor s/\eta \rfloor\eta})ds+\int_0^t (\Si_s-\widetilde \Si^{\eta}_{\lfloor s/\eta \rfloor\eta})d\bW_s
	\end{split}
\end{equation*}
Indeed, note that
\begin{equation*}
	\begin{split}
		\sup_{0\le t\le T}\|\bbeta_t-\widetilde \bbeta^{\eta}_t\|&\le \int_0^T\|\nabla G(\bbeta_s)-\nabla\widetilde G(\widetilde \bbeta^{\eta}_{\lfloor s/\eta \rfloor\eta})\|)ds+\sup_{0\le t\le T}\left\|\int_0^t (\Si_s-\widetilde \Si^{\eta}_{\lfloor s/\eta \rfloor\eta})d\bW_s\right\|
	\end{split}
\end{equation*}
We first square both sides and take expectation, then apply the Burkholder-Davis-Gundy inequality and  Cauchy-Schwarz inequality, we have 
\begin{equation}\label{sup norm estimate 1}
	\begin{split}
	\hE[	\sup_{0\le t\le T}\|\bbeta_t-\widetilde \bbeta^{\eta}_t\|^2]&\le 2\hE\left[\left(\int_0^T\|\nabla G(\bbeta_s)-\nabla\widetilde G(\widetilde \bbeta^{\eta}_{\lfloor s/\eta \rfloor\eta})\|ds\right)^2+\sup_{0\le t\le T}\left\|\int_0^t (\Si_s-\widetilde \Si^{\eta}_{\lfloor s/\eta \rfloor\eta})d\bW_s\right\|^2\right]\\
	&\le \underbrace{2T\hE\left[\int_0^T\|\nabla G(\bbeta_s)-\nabla\widetilde G(\widetilde \bbeta^{\eta}_{\lfloor s/\eta \rfloor\eta})\|^2ds \right]}_{\cI}+\underbrace{8\hE\left[\int_0^T\|\Si_s-\widetilde \Si^{\eta}_{\lfloor s/\eta \rfloor\eta}\|^2 ds\right]}_{\cJ}
	\end{split}
\end{equation}

\noindent
$\textbf{Estimate of stochastic gradient:}$
For the first term $\cI$, by using the inequality
$$\|a+b+c\|^2\le 3(\|a\|^2+\|b\|^2+\|c\|^2),$$ 
we get  
\begin{equation}
	\begin{split}
		\cI=&2T\hE\left[\int_0^T\left\|\left(\nabla G(\bbeta_s)-\nabla G(\widetilde \bbeta^{\eta}_s)\right)+\left(\nabla G(\widetilde \bbeta^{\eta}_s)-\nabla  G(\widetilde \bbeta^{\eta}_{\lfloor s/\eta \rfloor\eta})\right)+\left(\nabla  G(\widetilde \bbeta^{\eta}_{\lfloor s/\eta \rfloor\eta})-\nabla \widetilde G(\widetilde \bbeta^{\eta}_{\lfloor s/\eta \rfloor\eta})\right)\right\|^2ds \right]\\
		\le& \underbrace{6T\hE\left[\int_0^T\|\nabla G(\bbeta_s)-\nabla  G(\widetilde \bbeta^{\eta}_s)\|^2ds\right]}_{\cI_1}+\underbrace{6T\hE\left[\int_0^T\|\nabla  G(\widetilde \bbeta^{\eta}_s)-\nabla  G(\widetilde \bbeta^{\eta}_{\lfloor s/\eta \rfloor\eta})\|^2ds\right]}_{\cI_2}\\
		&+\underbrace{6T\hE\left[\int_0^T\|\nabla  G(\widetilde \bbeta^{\eta}_{\lfloor s/\eta \rfloor\eta})-\nabla \widetilde G(\widetilde \bbeta^{\eta}_{\lfloor s/\eta \rfloor\eta})\|^2ds\right]}_{\cI_3}\\
		\le& \cI_1+\cI_2+\cI_3.
	\end{split}
\end{equation}
By using the smoothness assumption \ref{assump: lip and alpha beta}, we first estimate 
\begin{equation*}
    \cI_1 \le 6TC^2\hE\left[\int_0^T\|\bbeta_s-\widetilde \bbeta^{\eta}_s\|^2ds\right].
\end{equation*}{}
By applying the smoothness assumption \ref{assump: lip and alpha beta} and discretization scheme, we can further estimate 
\begin{equation}\label{est cI2}
    \begin{split}
        \cI_2&\le 6TC^2\hE\left[\int_0^T\|\widetilde \bbeta^{\eta}_s-\widetilde \bbeta^{\eta}_{\lfloor s/\eta \rfloor\eta}\|^2ds\right]\\
        &\le 6TC^2\sum_{k=0}^{\lfloor T/\eta\rfloor} \hE\left[\int_{k\eta}^{(k+1)\eta}\|\widetilde \bbeta^{\eta}_s-\widetilde \bbeta^{\eta}_{\lfloor s/\eta \rfloor\eta} \|^2ds \right]\\
        &\le 6TC^2\sum_{k=0}^{\lfloor T/\eta\rfloor} \int_{k\eta}^{(k+1)\eta}\hE\left[\sup_{k\eta\le s<(k+1)\eta}\|\widetilde \bbeta^{\eta}_s-\widetilde \bbeta^{\eta}_{\lfloor s/\eta \rfloor\eta} \|^2\right]ds 
    \end{split}{}
\end{equation}{}
For $\forall~ k\in\mathbb N$ and $s\in [k\eta,(k+1)\eta)$, we have
\begin{equation*}
    \begin{split}
        \widetilde \bbeta^{\eta}_s-\widetilde \bbeta^{\eta}_{\lfloor s/\eta \rfloor\eta}=\widetilde \bbeta^{\eta}_s-\widetilde \bbeta^{\eta}_{k\eta}=-\nabla\widetilde G(\widetilde \bbeta^{\eta}_{k\eta})\cdot(s-k\eta)+\widetilde\Si^{\eta}_{k\eta}\int_{k\eta}^sd\bW_r
    \end{split}{}
\end{equation*}{}
which indeed implies
\begin{equation*}
    \begin{split}
     \sup_{k\eta\le s<(k+1)\eta}  \| \widetilde \bbeta^{\eta}_s-\widetilde \bbeta^{\eta}_{\lfloor s/\eta \rfloor\eta}\|\le \|\nabla\widetilde G(\widetilde \bbeta^{\eta}_{k\eta})\|(s-k\eta)+\sup_{k\eta\le s<(k+1)\eta} \|\widetilde\Si^{\eta}_{k\eta}\int_{k\eta}^sd\bW_r\|
    \end{split}{}
\end{equation*}{}
Similar to the estimate \eqref{sup norm estimate 1}, square both sides and take expectation, then apply the Burkholder-Davis-Gundy inequality, we have
\begin{equation*}
    \begin{split}
        \hE\left[\sup_{k\eta\le s<(k+1)\eta}  \| \widetilde \bbeta^{\eta}_s-\widetilde \bbeta^{\eta}_{\lfloor s/\eta \rfloor\eta}\|^2 \right]&\le 2\hE[ \|\nabla\widetilde G(\widetilde \bbeta^{\eta}_{k\eta})\|^2(s-k\eta)^2 ]+8\sum_{j=1}^{2d}\hE\left[\left(\widetilde \Si^{\eta}_{k\eta}(j)\la \int_{k\eta}^{\cdot}d\bW_r \ra_s^{1/2}\right)^2 \right]\\
         &\le  2(s-k\eta)^2\hE[ \|\nabla \widetilde G(\widetilde \bbeta^{\eta}_{k\eta})\|^2]+32d\tau_2(s-k\eta),
    \end{split}{}
\end{equation*}{}
where the last inequality follows from the fact that $\widetilde \Si^{\eta}_{k\eta}$ is a diagonal matrix with diagonal elements $\sqrt{2\tau_1}$ or $\sqrt{2\tau_2}.$
For the first term in the above inequality, we further have
\begin{equation*}
    \begin{split}
  2(s-k\eta)^2 \hE[ \|\nabla \widetilde G(\widetilde \bbeta^{\eta}_{k\eta})\|^2]&=        2(s-k\eta)^2\hE[ \|(\nabla G(\widetilde \bbeta^{\eta}_{k\eta})+\bphi_k)\|^2 ]\\
  &\le 4\eta^2\hE[ \|\nabla  G(\widetilde \bbeta^{\eta}_{k\eta})-\nabla  G(\bbeta^*) \|^2+\|\bphi_k\|^2 ]\\
   &\le 8C^2\eta^2\hE[\|\widetilde \bbeta^{\eta}_{k\eta}\|^2+\|\bbeta^*\|^2]+4\eta^2\hE[\|\bphi_k\|^2 ],
    \end{split}{}
\end{equation*}
where the first inequality follows from the separation of the noise from the stochastic gradient and the choice of stationary point $\bbeta^*$ of $G(\cdot)$ with $\nabla G(\bbeta^*)=0$, and $\bphi_k$ is the stochastic noise in the gradient at step $k$. Thus, combining the above two parts and integrate $ \hE\left[\sup_{k\eta\le s<(k+1)\eta}  \| \widetilde \bbeta^{\eta}_s-\widetilde \bbeta^{\eta}_{k\eta}\|^2 \right]$ on the time interval $[k\eta,(k+1)\eta)$, we obtain the following bound 
\begin{equation}\label{est cI2 part two}
    \begin{split}
        \int_{k\eta}^{(k+1)\eta}  \hE\left[\sup_{k\eta\le s<(k+1)\eta}  \| \widetilde \bbeta^{\eta}_s-\widetilde \bbeta^{\eta}_{k\eta}\|^2 \right] ds&\le 8C^2\eta^3 \left(\sup_{k\ge 0}\hE[ \|\widetilde \bbeta_{k\eta}^{\eta}\|^2+ \|\bbeta^*\|^2] \right)+4\eta^3 \max_{k}\hE[\|\bphi_k\|^2]+32d\tau_2\eta^2
    \end{split}{}
\end{equation}{}
By plugging the estimate \eqref{est cI2 part two} into estimate \eqref{est cI2}, we obtain the following estimates when $\eta\le 1$,
\begin{equation}
    \begin{split}
        \cI_2&\le 6TC^2(1+T/\eta)\left[8C^2\eta^3 \left(\sup_{k\ge 0}\hE[ \|\widetilde \bbeta_{k\eta}^{\eta}\|^2+\|\bbeta^*\|^2] \right)+4\eta^3 \max_{k}\hE[\|\bphi_k\|^2]+32d\tau_2\eta^2\right]\\
        & \le \tilde \delta_1(d, \tau_2, T, C, a, b) \eta + 24TC^2(1+T)\max_{k}\hE[\|\bphi_k\|^2],
    \end{split}{}
\end{equation}{}
where $\tilde \delta_1(d, \tau_2, T, C, a, b)$ is a constant depending on $d, \tau_2, T, C, a$ and $b$. Note that the above inequality requires a result on the bounded second moment of   $\sup_{k\ge 0}\hE[ \|\widetilde \bbeta_{k\eta}^{\eta}\|^2]$, and this is majorly\footnote{The slight difference is that the constant in the RHS of (C.38) \citet{chen2018accelerating} is changed to account for the stochastic noise.} proved in Lemma $C.2$ in \citet{chen2018accelerating} when we choose the stepzise $\eta\in (0, a/C^2)$. We are now left to estimate the term $\cI_3$ and we have
\begin{equation}
    \begin{split}
   \cI_3  &\le  6T\sum_{k=0}^{\lfloor T/\eta \rfloor} \hE\left[\int_{k\eta}^{(k+1)\eta}\|\nabla  G(\widetilde \bbeta^{\eta}_{k\eta})-\nabla \widetilde G(\widetilde \bbeta^{\eta}_{k\eta})\|^2ds \right]\\
   &\le 6T (1+T/\eta)\max_{k}\hE[ \|\bphi_k\|^2]\eta\\
   &\le 6T (1+T)\max_{k}\hE[ \|\bphi_k\|^2].
    \end{split}{}
\end{equation}{}
Combing all the estimates of $\cI_1,\cI_2$ and $\cI_3$, we obtain
\begin{equation}
    \begin{split}
        \cI\le& \underbrace{6TC^2\int_0^T\hE\left[\sup_{0\le s\le T}\|\bbeta_s-\widetilde \bbeta^{\eta}_s\|^2\right]ds}_{\cI_1}+\underbrace{\tilde \delta_1(d, \tau_2, T, C, a, b) \eta+ 24TC^2(1+T)\max_{k}\hE[\|\bphi_k\|^2]}_{\cI_2}\\
        &+\underbrace{6T (1+T)\max_{k}\hE[ \|\bphi_k\|^2]}_{\cI_3}.\\
    \end{split}{}
\end{equation}{}
\noindent
$\textbf{Estimate of stochastic diffusion:}$
For the second term $\cJ$, we have 
\begin{equation}
	\begin{split}
		\cJ&=8\hE\left[\int_{0}^{T} \|\Sigma_s(j)-\widetilde \Sigma_{{\lfloor s/\eta \rfloor}\eta}(j)\|^2 ds\right] \\
		& \leq 8\sum_{j=1}^{2d}\sum_{k=0}^{{\lfloor T/\eta \rfloor}}\int_{k \eta}^{(k+1)\eta}\hE\left[\|\Sigma_s(j)-\widetilde \Sigma^{\eta}_{k \eta}(j)\|^2\right]ds \\
		& \leq 8\sum_{j=1}^{2d}\sum_{k=0}^{{\lfloor T/\eta \rfloor}}\int_{k \eta}^{(k+1)\eta}\hE\left[\|\Sigma_s(j)- \Sigma_{k \eta}^{\eta}(j)+\Sigma_{k \eta}^{\eta}(j)-\widetilde \Sigma_{k \eta}^{\eta}(j)\|^2\right]ds \\
		& \leq 16\sum_{j=1}^{2d}\sum_{k=0}^{{\lfloor T/\eta \rfloor}}\left[\underbrace{\int_{k \eta}^{(k+1)\eta}\hE\left[\|\Sigma_s(j)- \Sigma_{k \eta}^{\eta}(j)\|^2\right]ds}_{\cJ_1}+\underbrace{\int_{k \eta}^{(k+1)\eta}\hE\left[\|\Sigma_{k \eta}^{\eta}(j)-\widetilde \Sigma_{k \eta}^{\eta}(j)\|^2\right]ds}_{\cJ_2}\right] 
	\end{split}.
\end{equation}
where $\Si_{k \eta}^{\eta}$ is the temperature matrix for the continuous-time interpolation of $\{\bbeta^{\eta}(k)\}_{k\ge 1}$, which is similar to \eqref{SGD continuous time interpolation} without noise generated from mini-batch settings and is defined as below
\begin{equation}
	 \bbeta_t^{\eta}= \bbeta_0-\int_0^t\nabla  G( \bbeta^{\eta}_{k\eta})ds+\int_0^t\Si^{\eta}_{k \eta}d\bW_s.
\end{equation}

We estimate $\cJ_1$ first, considering that $\Si_s$ and $\Si^{\eta}_{\lfloor s/\eta\rfloor \eta}$ are both diagonal matrices, we have
\begin{equation*}
    \begin{split}
        \cJ_1&=4(\sqrt{\tau_2}-\sqrt{\tau_1})^2\int_{k \eta}^{(k+1)\eta}\hP(\Si_s(j)\neq \Si^{\eta}_{k \eta}(j))ds\\
        &=4(\sqrt{\tau_2}-\sqrt{\tau_1})^2\hE\left[\int_{k \eta}^{(k+1)\eta}\hP(\Si_s(j)\neq \Si^{\eta}_{k \eta}(j)|\bbeta^{\eta}_{k \eta})ds\right]\\
        &= 4(\sqrt{\tau_2}-\sqrt{\tau_1})^2 r\int_{k \eta}^{(k+1)\eta} [(s-k \eta)+\mathcal R(s-k \eta)]ds\\
        &\le \tilde \delta_2(r, \tau_1,\tau_2)\eta^2,
    \end{split}{}
\end{equation*}{}
where $\tilde \delta_2(r, \tau_1,\tau_2)=4(\sqrt{\tau_2}-\sqrt{\tau_1})^2 r$, and the equality follows from the fact that the conditional probability $\hP(\Si_s(j)\neq \Si^{\eta}_{k \eta}(j)|\bbeta^{\eta}_{k \eta})=r S(\bbeta^{\eta(1)}_{k \eta},\bbeta^{\eta(2)}_{k \eta})\cdot(s-\eta)+r\mathcal R(s-k \eta)$. Here $\mathcal R(s-k \eta)$ denotes the higher remainder with respect to $s-k \eta$. The estimate of $\cJ_1$ without stochastic gradient for the Langevin diffusion is first obtained in \citet{chen2018accelerating}, we however present here again for reader's convenience.  
As for the second term $\cJ_2$, it follows that
\begin{equation}
	\begin{split}
	\label{new_prob}
	    \cJ_2&=4(\sqrt{\tau_2}-\sqrt{\tau_1})^2\int_{k \eta}^{(k+1)\eta}\hP(\Sigma_{k \eta}(j)\neq \widetilde \Sigma_{k \eta}(j))ds \\
	    &= 4(\sqrt{\tau_2}-\sqrt{\tau_1})^2 r\eta \hE\left[\left|S(\bbeta_{k \eta}^{\eta(1)}, \bbeta_{k \eta}^{\eta(2)})-\tilde S(\widetilde \bbeta_{k \eta}^{\eta(1)}, \widetilde \bbeta_{k \eta}^{\eta(2)})\right|\right] \\
	    &\le \tilde \delta_2(r, \tau_1,\tau_2) \eta \sqrt{\hE\left[\left|S(\bbeta_{k \eta}^{\eta(1)}, \bbeta_{k \eta}^{\eta(2)})-\tilde S(\widetilde \bbeta_{k \eta}^{\eta(1)}, \widetilde \bbeta_{k \eta}^{\eta(2)})\right|^2\right]}\\
	    & \leq \tilde \delta_2(r, \tau_1,\tau_2) \eta \sqrt{\hE\left[|\psi_{k}|^2\right]},
	\end{split}
\end{equation}
where $\psi_{k}$ is the noise in the swapping rate. Thus, one concludes the following estimates combing $\cI$ and $\cJ$.
\begin{equation}
	\begin{split}
	&\hE[\sup_{0\le t\le T}\|\bbeta_t-\widetilde \bbeta^{\eta}_t||^2] \le \underbrace{6TC^2\int_0^T\hE\left[\sup_{0\le s\le T}\|\bbeta_s-\widetilde \bbeta^{\eta}_s\|^2\right]ds}_{\cI_1}+\underbrace{\tilde \delta_1(d, \tau_2, T, C, a, b) \eta+ 24TC^2(\eta+T)\max_{k}\hE[\|\bphi_k\|^2]}_{\cI_2}\\
	&\ \ \ \ \ +\underbrace{6T (1+T) \hE[\|\bphi_{k}\|^2]}_{\cI_3}+\underbrace{32d(1+T)\tilde \delta_2(r, \tau_1,\tau_2)\left(\eta+\max_{k}\sqrt{\hE\left[|\psi_{k}|^2\right]}\right)}_{\cJ}. 
	\end{split}
\end{equation}
Apply Gronwall's inequality to the function
\begin{equation*}
    t\mapsto \hE\left[\sup_{0\le u\le t} \|\bbeta_u-\widetilde \bbeta_u^{\eta}\|^2 \right],
\end{equation*}{}
and deduce that 
\begin{equation}
	\begin{split}
&	\hE[\sup_{0\le t\le T}\|\bbeta_t-\widetilde \bbeta^{\eta}_t||^2] \le D_1 \eta + D_2 \max_{k}\hE[\|\bphi_k\|^2]+D_3\max_{k}\sqrt{\hE\left[|\psi_{k}|^2\right]},\\
	\end{split}
\end{equation}
where $ D_1$ is a constant depending on $\tau_1,\tau_2,d, T, C,a,b$; $D_2$ depends on $T$ and $C$; $D_3$ depends on $r, d, T$ and $C$. \qed

\end{proof}

\subsection{Exponential decay of Wasserstein distance in continuous-time}

We proceed to quantify the evolution of the 2-Wasserstein distance between $\n_{t}$ and $\pi$. We first consider the ordinary Langevin diffusion without swaps and derive the log-Sobolev inequality (LSI). Then we extend LSI to reLD and obtain the exponential decay of the relative entropy. Finally, we derive the exponential decay of the 2-Wasserstein distance.

In order to distinguish from the replica exchange Langevin diffusion $\bbeta_t$ defined in \eqref{replica exchange}, we call it $\hat \bbeta_t$ which follows, 
\begin{equation}
    d\hat \bbeta_t=-\nabla G(\hat \bbeta_t)dt+\Si_td\bW_t.
\end{equation}{}
where $\Si_t\in \hR^{2d\times 2d}$ is a diagonal matrix with the form $\begin{pmatrix}{}
\sqrt{2\tau_1}\mathbf I_d&0\\
0&\sqrt{2\tau_2}\mathbf I_d
\end{pmatrix}$. 
The process $\hat \bbeta_t$ is a Markov diffusion process with infinitesimal generator $\cL$ in the following form, for $x_1\in\hR^d$ and $x_2\in\hR^d$,
\beaa
\cL=&-\la\nabla_{x_1}f(x_1,x_2),\nabla U(x_1)\ra+\tau_1\Delta_{x_1}f(x_1,x_2)\\
&-\la \nabla_{x_2}f(x_1,x_2),\nabla U(x_2)\ra+\tau_2\Delta_{x_2}f(x_1,x_2)
\eeaa
Note that since matrix $\Si_t$ is a non-degenerate diagonal matrix, operator $\cL$ is an elliptic diffusion operator. According to the smoothness assumption \eqref{assump: lip and alpha beta}, we have that $\nabla^2 G\ge -C\mathbf I_{2d}$, where $C>0$, the unique invariant measure $\pi$ associate with the underlying diffusion process satisfies the Poincare inequality and LSI with the Dirichlet form given as follows,
\bea\label{dirichlet form}
\cE(f)=\int \Big(\tau_1\|\nabla_{x_1}f\|^2+\tau_2\|\nabla_{x_2}f\|^2 \Big)d\pi(x_1,x_2),\qq f\in\cC_0^2(\hR^{2d}).
\eea
In this elliptic case with $G$ being convex, the proof for LSI follows from standard Bakry-Emery calculus \cite{Bakry85}. Since, we are dealing with the non-convex function $G$, we are particularly interested in the case of $\nabla^2 G\ge -C\mathbf I_{2d}$.
To obtain a Poincar\'{e} inequality for invariant measure $\pi$, \citet{chen2018accelerating} adapted an argument from \citet{Bakry08} and \citet{Maxim17} by constructing an appropriate Lyapunov function for the replica exchange diffusion without swapping $\hat \bbeta_t$. Denote $\n_t$ as the distribution associated with the diffusion process $\{\hat \bbeta_t\}_{t\ge 0}$, which is absolutely continuous with respect to $\pi$. It is a direct consequence of the aforementioned results that the following log-Sobolev inequality holds.

\begin{lemma}[LSI for Langevin Diffusion]\label{LSI no swaping}
Under assumptions \eqref{assump: lip and alpha beta} and \eqref{assump: dissipitive}, we have the following log-Sobolev inequality for invariant measure $\pi$, for some constant $c_{\text{LS}}>0$, 
\beaa
D(\n_t||\pi)\le 2c_{\text{LS}}\cE(\sqrt{\frac{d\n_t}{d\pi}}).
\eeaa
where $D(\n_t||\pi)=\int d\nu_t \log\frac{d\nu_t}{d\pi}$ denotes the relative entropy and the Dirichlet form $\cE(\cd)$ is defined in \eqref{dirichlet form}.
\end{lemma}{}
\begin{proof}

According to \citet{Cattiaux2010}, the sufficient conditions to establish LSI are:
\begin{enumerate}
\item There exists some constant $C\ge 0$, such that $\nabla^2 G\succcurlyeq -C I_{2d}$.
\item $\pi$ satisfies a Poincar\'{e} inequality with constant $c_{p}$, namely, for all probability measures $\nu\ll\pi$, $\chi^2(\nu||\pi)\leq c_p \cE(\sqrt{\frac{d\n_t}{d\pi}})$, where $\chi^2(\nu||\pi):=\|\frac{d\nu}{d\pi}-1\|^2$ is the $\chi^2$ divergence between $\nu$ and $\pi$.
\item There exists a $\cC^2$ Lyapunov function $V: \mathbb{R}^{2d}\rightarrow [1, \infty)$ such that $\frac{\cL V(x_1, x_2)}{V(x_1, x_2)} \leq \kappa - \gamma (\|x_1\|^2 + \|x_2\|^2)$
for all $(x_1, x_2)\in \mathbb{R}^{2d}$ and some $\kappa, \gamma>0$.
\end{enumerate}
Note that the first condition on the Hessian is obtained from the smoothness assumption \eqref{assump: lip and alpha beta}. Moreover, the Poincar\'{e} inequality in the second condition is derived from Lemma C.1 in \citet{chen2018accelerating} given assumptions \eqref{assump: lip and alpha beta} and \eqref{assump: dissipitive}. Finally, to verify the third condition, we follow \citet{Maxim17} and construct the Lyapunov function 
$V(x_1,x_2):=\exp\left\{a/4 \cdot \left(\frac{\|x_1\|^2}{\tau_1}+\frac{\|x_2\|^2}{\tau_2}\right)\right\}$. From the dissipitive assumption \ref{assump: dissipitive}, $V(x_1, x_2)$ satisfies the third condition because
\begin{equation}
\begin{split}
    \cL(V(x_1, x_2))&=\left(\frac{a}{2\tau_1}+\frac{a}{2\tau_2}+\frac{a^2}{4\tau_1^2}\|x_1\|^2+\frac{a^2}{4\tau_2^2}\|x_2\|^2-\frac{a}{2\tau_1^2}\langle x_1, \nabla G(x_1)-\frac{a}{2\tau_2^2}\langle x_1, \nabla G(x_2)\rangle\right) V(x_1, x_2)\\
    &\leq \left(\frac{a}{2\tau_1}+\frac{a}{2\tau_2}+\frac{ab}{2\tau_1^2}+\frac{ab}{2\tau_2^2}-\frac{a^2}{4\tau_1^2}\|x_1\|^2-\frac{a^2}{4\tau_2^2}\|x_2\|^2\right) V(x_1, x_2)\\
    &\leq \left(\kappa-\gamma (\|x_1\|^2+\|x_2\|^2)\right) V(x_1, x_2),\\
\end{split}
\end{equation}
where $\kappa=\frac{a}{2\tau_1}+\frac{a}{2\tau_2}+\frac{ab}{2\tau_1^2}+\frac{ab}{2\tau_2^2}$, and $\gamma=\frac{a^2}{4\tau_1^2}\land \frac{a^2}{4\tau_2^2}$. Therefore, the invariant measure $\pi$ satisfies a LSI with the constant
\begin{equation}
    c_{\text{LS}}=c_1+(c_2+2)c_p,
\end{equation}
where $c_1=\frac{2C}{\gamma}+\frac{2}{C}$ and $c_2=\frac{2C}{\gamma}\left(\kappa+\gamma\int_{\mathbb{R}^{2d}}( \|x_1\|^2 + \|x_2\|^2)\pi(dx_1 dx_2)\right)$. \qed


\end{proof}{}

We are now ready to prove the log-Sobolev inequality for invariant measure associated with the replica exchange Langevin diffusion \eqref{replica exchange}. We use a similar idea from \citet{chen2018accelerating} where they prove the Poincar\'{e} inequality for the invariant measure associated with the replica exchange Langevin diffusion \eqref{replica exchange} by analyzing the corresponding Dirichlet form. In particular, a larger Dirichlet form ensures a smaller log-Sobolev constant and hence results in a faster convergence in the relative entropy and Wasserstein distance.

\begin{lemma}[Accelerated exponential decay of $\mathcal{W}_2$]\label{exponential decay}
Under assumptions \eqref{assump: lip and alpha beta} and \eqref{assump: dissipitive}, we have that the replica exchange Langevin diffusion converges exponentially fast to the invariant distribution $\pi$:
\begin{equation}
    \mathcal{W}_2(\nu_t,\pi) \leq  D_0 e^{-k\eta(1+\delta_S)/c_{\text{LS}}},
\end{equation}
where $D_0=\sqrt{2c_{\text{LS}}D(\nu_0||\pi)}$, $\delta_{S}:=\inf_{t>0}\frac{\cE_S(\sqrt{\frac{d\n_t}{d\pi}})}{\cE(\sqrt{\frac{d\n_t}{d\pi}})}-1$ is a non-negative constant depending on the swapping rate $S(\cd, \cd)$ and obtains $0$ only if $S(\cd, \cd)=0$.
\end{lemma}{}

\begin{proof}
Consider the infinitesimal generator associated with the diffusion process \eqref{replica exchange}, denoted as $\cL_{S}$, contains an extra term arising from the temperature swapping. The operator $\cL_{S}$ in this particular case, indeed, has the following form
\bea
\cL_{S}=\cL+ rS(x_1,x_2)\cd (f(x_2,x_1)-f(x_1,x_2)).
\eea
According to Theorem 3.3 \citep{chen2018accelerating}, the Dirichlet form associated with operator $\cL_{S}$ under the invariant measure $\pi$ has the form
\bea\label{dirichlet swap}
\cE_{S}(f)=\cE(f)+\underbrace{\frac{r}{2}\int S(x_1,x_2)\cd (f(x_2,x_1)-f(x_1,x_2))^2d\pi(x_1,x_2)}_{\text{acceleration}},~ f\in\cC_0^2(\hR^{2d}),
\eea
where $f$ corresponds to $\frac{d\nu_t}{d\pi(x_1, x_2)}$, and the asymmetry of $\frac{\nu_t}{\pi(x_1, x_2)}$ is critical in the acceleration effect \citep{chen2018accelerating}. Given two different temperatures $\tau_1$ and $\tau_2$, a non-trivial distribution $\pi$ and function $f$, the swapping rate $S(x_1,x_2)$ is positive for almost any $x_1, x_2\in\mathbb{R}^d$. As a result, the Dirichlet form associated with $\cL_{S}$ is strictly larger than $\cL$. Therefore, there exists a constant $\delta_{S}> 0$ depending on $S(x_1,x_2)$, such that $\delta_{S}=\inf_{t>0}\frac{\cE_S(\sqrt{\frac{d\n_t}{d\pi}})}{\cE(\sqrt{\frac{d\n_t}{d\pi}})}-1$. From Lemma \ref{LSI no swaping}, we have
\begin{equation}
    D(\n_t||\pi)\le 2c_{\text{LS}}\cE(\sqrt{\frac{d\n_t}{d\pi}})\le 2c_{\text{LS}} \sup_t\frac{ \cE(\sqrt{\frac{d\n_t}{d\pi}})}{\cE_{S}(\sqrt{\frac{d\n_t}{d\pi}})}\cE_{S}(\sqrt{\frac{d\n_t}{d\pi}})= 2 \frac{c_{\text{LS}}}{1+\delta_{S}}\cE_{S}(\sqrt{\frac{d\n_t}{d\pi}}).
\end{equation}
Thus, we obtain the following log-Sobolev inequality for the unique invariant measure $\pi$  associated with replica exchange Langevin diffusion $\{\bbeta_t\}_{t\ge 0}$ and its corresponding Dirichlet form $\cE_{S}(\cd)$. In particular, the LSI constant $ \frac{c_{\text{LS}}}{1+\delta_{S}}$ in replica exchange Langevin diffusion with swapping rate $S(\cd, \cd)>0$ is strictly smaller than the LSI constant $c_{\text{LS}}$ in the replica exchange Langevin diffusion with swapping rate $S(\cd, \cd)=0$. By the exponential decay in entropy \citep{Bakry2014}[Theorem 5.2.1] and the tight log-Sobolev inequality in Lemma \ref{LSI no swaping}, we get that, for any $t\in[k\eta,(k+1)\eta)$, 
\begin{equation}
    D(\nu_t||\pi)\leq D(\nu_0||\pi) e^{-2t(1+\delta_S)/c_{\text{LS}}}\leq D(\m_0||\pi) e^{-2k\eta(1+\delta_S)/c_{\text{LS}}}.
\end{equation}
Finally, we can estimate the term $\mathcal{W}_2(\nu_t,\pi)$ by the
Otto-Villani theorem \citep{Bakry2014}[Theorem 9.6.1],
\begin{equation}
    \mathcal{W}_2(\nu_t,\pi) \leq \sqrt{2 c_{\text{LS}} D(\nu_t||\pi)}\leq \sqrt{2c_{\text{LS}}D(\m_0||\pi)} e^{-k\eta(1+\delta_S)/c_{\text{LS}}}.
\end{equation}
\end{proof}


\subsection{Summary: Convergence of reSGLD}

Now that we have all the necessary ingredients in place, we are ready to derive the convergence of the distribution $\m_{k}$ to the invariant measure $\pi$ in terms of 2-Wasserstein distance, 


\begin{theorem}[Convergence of reSGLD]Let the assumptions \eqref{assump: lip and alpha beta} and \eqref{assump: dissipitive} hold. For the unique invariant measure $\pi$ associated with the Markov diffusion process \eqref{replica exchange} and the distribution $\{\m_{k}\}_{k\ge 0}$ associated with the discrete dynamics $\{\widetilde \bbeta^{\eta}(k)\}_{k\ge 1}$, we have the following estimates, for $0\le k\in \mathbb N^{+}$ and the learning rate $\eta$ satisfying $0<\eta<1 \land a/C^2$, 
\begin{equation}
    \mathcal{W}_2(\m_{k}, \pi) \le  D_0 e^{-k\eta(1+\delta_S)/c_{\text{LS}}}+\sqrt{\delta_1 \eta +  \delta_2 \max_{k}\hE[\|\bphi_k\|^2]+ \delta_3\max_{k}\sqrt{\hE\left[|\psi_{k}|^2\right]}}
\end{equation}
where $D_0=\sqrt{2c_{\text{LS}}D(\m_0||\pi)}$, $\delta_{S}:=\min_{k}\frac{\cE_S(\sqrt{\frac{d\m_k}{d\pi}})}{\cE(\sqrt{\frac{d\m_k}{d\pi}})}-1$ is a non-negative constant depending on the swapping rate $S(\cd, \cd)$ and obtains the minimum zero only if $S(\cd, \cd)=0$.
\end{theorem}{}

\begin{proof}
We reduce the estimates into the following two terms by using the triangle inequality,
\begin{equation}\label{w2 triangle}
    \mathcal{W}_2(\m_{k}, \pi) \leq \mathcal{W}_2(\m_{k}, \n_t) + \mathcal{W}_2(\nu_t,\pi),\qq t\in[k\eta,(k+1)\eta).
\end{equation}
The first term $\mathcal{W}_2(\m_{k}, \n_t)$ follows from the analysis of discretization error in Lemma.\ref{discretization}.
Recall the very definition of the $\mathcal{W}_2(\cd,\cd)$ distance defined in (\ref{w2}). Thus, in order to control the distance $\mathcal{W}_2(\m_{k},\nu_t)$, $t\in[k\eta,(k+1)\eta)$, we need to consider the diffusion process whose law give $\m_{k}$ and $\n_t$, respectively. Indeed, it is obvious that $\n_t=\cL(\bbeta_t)$ for $t\in[k\eta,(k+1)\eta)$. For the other measure $\m_k$, it follows that $\m_{k}=\tilde \n_{k\eta}$ for $t=k\eta$, where $\tilde\n_{k\eta}=\cL(\widetilde \bbeta_t^{\eta})$ is the probability measure associated with the continuous interpolation of reSGLD (\ref{resgld_2}). By Lemma.\ref{discretization}, we have that for $k\in\mathbb{N}$ and $ t\in [k\eta,(k+1)\eta)$, 
\begin{equation}
    \mathcal{W}_2(\m_{k}, \n_t)=\mathcal{W}_2(\tilde\n_{k\eta}, \n_t) \leq \sqrt{\hE[\sup_{0\le s\le t}\|\bbeta_s- \widetilde \bbeta_s^{\eta}\|^2]}\leq \sqrt{\delta_1 \eta + \delta_2 \max_{k}\hE[\|\bphi_k\|^2]+\delta_3\max_{k}\sqrt{\hE\left[|\psi_{k}|^2\right]}},
\end{equation}

Recall from the accelerated exponential decay of replica exchange Langevin diffusion in Lemma.\ref{exponential decay}, we have
\begin{equation}
    \mathcal{W}_2(\nu_t,\pi)\leq \sqrt{2c_{\text{LS}}D(\nu_0||\pi)} e^{-k\eta(1+\delta_S)/c_{\text{LS}}}= \sqrt{2c_{\text{LS}}D(\m_0||\pi)} e^{-k\eta(1+\delta_S)/c_{\text{LS}}}.
\end{equation}

Combing the above two estimates completes the proof.
\qed
\end{proof}{}

%% file: supp_B.tex
In the semi-supervised learning tasks, we fine-tune the hyper-parameters for Bayesian GANs and report them in Table \ref{hyper}. In particular, $N_s$ is the number of labeled data; $\eta^{(1)}$ and $\eta^{(2)}$ are the learning rates for the low-temperature chain and high-temperature chain, respectively; $\tau_1$ and $\tau_2$ are the temperatures; $\hat F$ is the correction factor, which often yields several swaps. In addition, the learning rates also follow a truncated exponential decay, for example, $\eta^{(1)}_k=\left(0.05 \vee e^{-\frac{k}{800}}\right)\eta^{(1)}$ and $\eta^{(2)}_k=\left(0.05 \vee e^{-\frac{k}{800}}\right)\eta^{(2)}$, where $k$ is the number of iterations.
\begin{table*}[ht]
  \centering
  \small
  \begin{tabular}{c|ccccccc}
    \toprule
    Dataset &   $N_s$ & $\eta^{(1)}$ &  $\eta^{(2)}$ & $\tau_1$ & $\tau_2$ & $\hat F$ \\
     \midrule
     \midrule
    \multirow{2}{*}{CIFAR10}  & 2000 $\sim$ 3500 & 4.5e-4 & 7.0e-4 & 0.01 & 1 & 3.0e5 \\
     & 4000 $\sim$ 5000 & 4.5e-4 & 7.0e-4 & 0.01 & 1 & 2.0e5 \\
    \midrule
    \multirow{3}{*}{CIFAR100}  & 2000 $\sim$ 2500 & 5.0e-4 & 7.5e-4 & 0.04 & 1 & 1.0e4 \\
     & 3000 $\sim$ 3500 & 5.0e-4 & 7.5e-4 & 0.02 & 1 & 2.5e4 \\ 
     & 4000 $\sim$ 5000 & 5.0e-4 & 7.5e-4 & 0.01 & 1 & 5.0e4 \\ 
    \midrule
    \multirow{2}{*}{SVHN} & 2000 $\sim$ 4000 & 4.5e-3 & 5.0e-3 & 0.01 & 1 & 8.0e4 \\
     & 4500 $\sim$ 5000 & 4.5e-3 & 7.0e-3 & 0.01 & 1 & 8.0e4 \\ 
    \bottomrule
  \end{tabular}
  \caption{Hyper-parameter setting of Bayesian GANs for Semi-Supervised Learning experiments. 
  }
  \label{hyper}
\end{table*}